\documentclass{article}

\usepackage{microtype}      
\usepackage[utf8]{inputenc} %
\usepackage[T1]{fontenc}    
\usepackage{tabularx}
\usepackage[table, dvipsnames]{xcolor} 
\usepackage{graphicx}
\usepackage{subcaption}     
\usepackage{float}          

\definecolor{MethodBlue}{RGB}{0, 70, 140}       
\definecolor{MethodBlueBg}{RGB}{240, 245, 250}   
\definecolor{RiskRed}{RGB}{180, 50, 50}         
\definecolor{GrayBg}{RGB}{245, 245, 245}        
\definecolor{DarkGreen}{RGB}{0, 150, 0}
\definecolor{myblue}{RGB}{0, 70, 140} 
\definecolor{BestRed}{RGB}{190, 0, 0} 
\newcommand{\best}[1]{\textcolor{BestRed}{\textbf{#1}}}

\usepackage[preprint]{icml2026} 

\usepackage{amsmath}
\usepackage{amssymb}
\usepackage{mathtools}
\usepackage{amsthm}

\usepackage{booktabs}       
\usepackage{multirow}       
\usepackage{enumitem}       

\usepackage{listings}       
\usepackage{tcolorbox}      
\tcbuselibrary{breakable}
\usepackage{hyperref}       
\hypersetup{
    colorlinks=true,
    linkcolor=myblue,
    filecolor=magenta,      
    urlcolor=myblue,
    citecolor=myblue,       
}

\usepackage[capitalize,noabbrev]{cleveref} 
\newcommand{\sd}[1]{{\scriptsize $\pm$#1}}
\usepackage[textsize=tiny]{todonotes} 

\theoremstyle{plain}

\theoremstyle{definition}

\theoremstyle{remark}

\icmltitlerunning{\textsc{ARGOS}: Automated Functional Safety Requirement Synthesis for Embodied AI via Attribute-Guided Combinatorial Reasoning}
\begin{document}

\twocolumn[

\icmltitle{\textsc{ARGOS}: Automated Functional Safety Requirement Synthesis for Embodied AI via Attribute-Guided Combinatorial Reasoning}



\icmlsetsymbol{equal}{*}

\begin{icmlauthorlist}
\icmlauthor{Dongsheng Chen}{sustech}
\icmlauthor{Yuxuan Li}{sustech}
\icmlauthor{Yi Lin}{sustech}
\icmlauthor{Guanhua Chen}{sustech}
\icmlauthor{Jiaxin Zhang}{sustech}
\icmlauthor{Xiangyu Zhao}{cityu}
\icmlauthor{Lei Ma}{utokyo}
\icmlauthor{Xin Yao}{lingnan}
\icmlauthor{Xuetao Wei}{sustech}
\end{icmlauthorlist}
\icmlaffiliation{sustech}{Southern University of Science and Technology, Shenzhen, China}
\icmlaffiliation{cityu}{City University of Hong Kong, Hong Kong, China}
\icmlaffiliation{utokyo}{The University of Tokyo, Tokyo, Japan}
\icmlaffiliation{lingnan}{Lingnan University, Hong Kong, China}

\icmlcorrespondingauthor{Xuetao Wei}{weixt@sustech.edu.cn}

\icmlkeywords{Machine Learning, ICML}

\vskip 0.3in
]



\printAffiliationsAndNotice{\icmlEqualContribution} 
\begin{abstract}


Ensuring functional safety is essential for the deployment of Embodied AI in complex open-world environments. 
However, traditional Hazard Analysis and Risk Assessment (HARA) methods struggle to scale in this domain. 
While HARA relies on enumerating risks for finite and pre-defined function lists, Embodied AI operates on open-ended natural language instructions, creating a challenge of combinatorial interaction risks. 
Whereas Large Language Models (LLMs) have emerged as a promising solution to this scalability challenge, they often lack physical grounding, yielding semantically superficial and incoherent hazard descriptions. 
To overcome these limitations, we propose a new framework \textbf{ARGOS} (\textbf{A}tt\textbf{R}ibute-\textbf{G}uided c\textbf{O}mbinatorial rea\textbf{S}oning), which bridges the gap between open-ended user instructions and concrete physical attributes. 
By dynamically decomposing entities from instructions into these fine-grained properties, ARGOS grounds LLM reasoning in causal risk factors to generate physically plausible hazard scenarios. 
It then instantiates abstract safety standards, such as ISO~13482, into context-specific Functional Safety Requirements (FSRs) by integrating these scenarios with robot capabilities. 
Extensive experiments validate that ARGOS produces high-quality FSRs and outperforms baselines in identifying long-tail risks. 
Overall, this work paves the way for systematic and grounded functional safety requirement generation, a critical step toward the safe industrial deployment of Embodied AI.
\end{abstract}

\begin{figure*}[t] 
    \centering
    \includegraphics[width=0.95\textwidth]{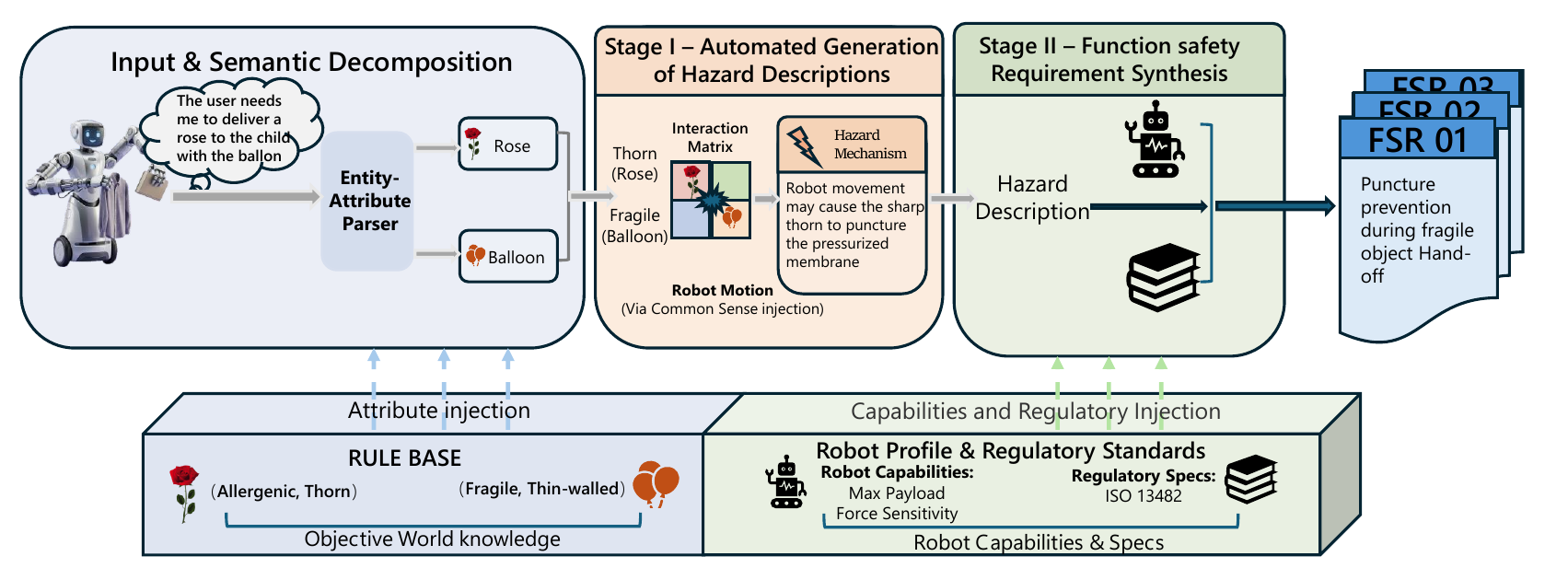} 
    \caption{The proposed ARGOS framework: A two-stage pipeline for automated FSR synthesis. Stage I focuses on decomposing semantic entities into physical attributes for combinatorial hazard discovery, while Stage II aligns these hazards with regulatory standards and hardware constraints to generate requirements.}
    \label{fig:overall_framework}
\end{figure*}
\section{Introduction}
The rapid evolution of Embodied AI is propelling robots from
constrained industrial environments into open household environments. While models like PaLM-E \cite{driess2023palm} and SayCan \cite{brohan2023can} enable robots to generalize across diverse tasks via natural language, this paradigm shift introduces unforeseen safety challenges.  Unlike traditional robotic systems that operate within a finite set of pre-defined functions, Embodied AI is driven by open-ended user instructions. This creates an  effectively infinite functional space , where robots must dynamically interpret arbitrary intents. Consequently, the unstructured nature of domestic settings, compounded by this open-ended task space, leads to a combinatorial explosion of task-environment interactions. As highlighted by Lasota et al. \cite{2017LasotaSurvey}, this results in a vast, non-enumerable risk space. This poses a fundamental 
scaling problem for traditional safety engineering: expert-based manual methods, such as Hazard Analysis and Risk Assessment (HARA), which inherently rely on enumerating risks for static, finite functional definitions are ill-equipped to handle this complexity. As Seshia et al. \cite{seshia2022toward} argue, \textbf{the contradiction between this infinite operational domain and the limited capacity of manual verification creates a critical bottleneck, necessitating automated, scalable mechanisms for risk discovery.}

To address these challenges, recent research \cite{shi2024aegis} \cite{nouri2024welcome} has explored using LLMs as ``expert surrogates'' to streamline functional safety requirements (FSR) generation. While efficient for standardizing documentation, these approaches primarily operate in structured engineering contexts and often overlook the dynamic physical nuances inherent in embodied interactions. We argue that the application of LLMs in these safety-critical domains faces a fundamental discrepancy between semantic processing and physical grounding. The root lies in the Symbol Grounding Problem \cite{bender2020climbing}: models trained solely on linguistic form cannot inherently grasp meaning without the ``shared experience'' of the physical world \cite{bisk2020experience}. Consequently, this lack of grounding leads to demonstrable physical incoherence. Valmeekam et al. \cite{valmeekam2022large} show that LLMs struggle with reasoning about state changes in dynamic environments, while the NEWTON benchmark \cite{wang2023newton} further reveals inconsistencies in object-attribute reasoning (e.g., material properties or mass), which is a critical limitation for identifying risks dependent on precise physical constraints.

Instead of learning causal mechanisms, LLMs often resort to shortcut learning \cite{geirhos2020shortcut}, relying on high-frequency statistical correlations (``Label-to-Label'' mapping) rather than first-principles deduction. As Kandpal et al. \cite{kandpal2023large} demonstrate, this statistical reliance results in a failure to capture long-tail knowledge, where the most dangerous, rare accident scenarios reside. Crucially, standard models often treat risk factors in isolation, failing to reason about the conflicting constraints introduced by concurrent tasks. For instance, while a model might flag a generic collision risk from a running child, it often fails to deduce the systemic hazard where the robot's necessary safety maneuver (e.g., an emergency stop) directly compromises the stability required by a delicate delivery task. By overlooking these coupled dynamics, implicit inference misses the dilemma-driven risks inherent in embodied interaction.

\textbf{To address these limitations, we propose a new framework ARGOS for automated functional safety requirement synthesis for Embodied AI via \textbf{A}tt\textbf{R}ibute-\textbf{G}uided c\textbf{O}mbinatorial rea\textbf{S}oning.} To tackle the challenge of infinite task possibilities, ARGOS moves beyond the traditional reliance on pre-defined function lists. Instead, it leverages the semantic generalization of LLMs to dynamically parse open-ended user instructions into structured risk factors. \textbf{This framework bridges the gap between the infinite semantic space of user commands and explicit physical constraints through a structured two-stage pipeline:}
\textbf{(1) Attribute-Guided Hazard Discovery:}
Utilizing a retrieve-and-infer mechanism, the system first extracts key risk factors from arbitrary natural-language instructions and injects their corresponding physical attributes.
It then guides the LLM to synthesize hazard scenarios by systematically simulating how these attributes interact with the seed scenario.
More generally, ARGOS supports \emph{$k$-factor} coupled risk reasoning by progressively exploring higher-order combinations of risk factors (we evaluate up to $k \le 3$ in our experiments for tractability). \textbf{(2) Scenario-Anchored Requirement Synthesis:} For the generated hazard scenarios, we perform a constrained synthesis to generate Testable FSRs. By explicitly conditioning the generation on a Robot Capability List and regulatory standards (e.g., ISO 13482), ARGOS ensures that the derived requirements are not only legally compliant but also physically feasible within the robot's hardware limits.

We focus on FSRs rather than Technical Safety Requirements (TSRs), as FSRs represent the optimal abstraction level for LLM-based reasoning. While LLMs excel at deriving high-level functional logic, synthesizing low-level technical parameters (TSRs) often leads to physical hallucinations. Importantly, FSRs provide the essential logical foundation; once the functional objectives are clearly defined, deriving specific TSRs becomes a systematic technical refinement within a well-defined design space.

We validated ARGOS on a curated dataset of 48 seed scenarios spanning diverse risk entities, tasks, and environments, which were expanded into 365 context-specific hazard scenarios via our pipeline.
In the Stage I evaluation, we compared ARGOS against Vanilla LLM and Physics-Aware Chain-of-Thought (CoT) baselines using three custom metrics. Both LLM-as-a-Judge and human evaluation demonstrate that ARGOS achieves superior performance across all indicators. Furthermore, Latent Semantic Topology analysis reveals that ARGOS effectively captures long-tail risks, exposing the semantic redundancy inherent in the CoT baseline.
In the Stage II evaluation, we conducted an ablation study using a similar multi-metric LLM-based assessment. The results confirm that the full ARGOS pipeline, integrating both attribute injection and regulatory constraints, achieves optimal performance in synthesizing high-quality FSRs compared to ablated variants.

\textbf{Our contributions are summarized as follows}:
\begin{itemize}[leftmargin=*, noitemsep, topsep=0pt, parsep=0pt]
\item To the best of our knowledge, we present the first automated framework capable of synthesizing FSRs directly from unconstrained, open-ended user instructions. This work addresses the critical scalability bottleneck of traditional HARA, enabling dynamic safety synthesis for the unbounded task spaces of modern Embodied AI.

\item  We propose a new framework ARGOS that shifts safety analysis from implicit ``Label-to-Label'' mapping to explicit Attribute-Based Deduction. By decomposing entities into physical constraints, ARGOS mitigates the symbol grounding problem, enabling the discovery of long-tail risks.

\item  Extensive experiments validate the efficacy of our framework, demonstrating high-quality in FSR generation. Furthermore, ARGOS outperforms baselines in discovering long-tail risks.

\end{itemize}

\section{Related Work}

\subsection{Safety Engineering in the Open World}
Safety assurance for personal care robots is governed by rigorous standards like ISO 13482 \cite{iso13482}. Traditionally, compliance relies on manual methodologies such as STPA \cite{leveson2016engineering}. However, the transition from structured industrial settings to unstructured household environments introduces the open-world challenge. Lasota et al. \cite{2017LasotaSurvey} and Seshia et al. \cite{seshia2022toward} argue that the unpredictability of human behavior creates a combinatorial explosion of risk scenarios, rendering static, expert-based enumeration insufficient. This creates a critical need for automated mechanisms capable of handling an effectively infinite operational design domain.

\subsection{From Software Requirements to Embodied AI}
In the broader software domain, LLMs have shown promise in automating requirements engineering. Reviews by Hou et al. \cite{hou2024large}, Fan et al. \cite{fan2023large}, and Marques et al. \cite{marques2024using} highlight their utility in elicitation and documentation. However, challenges remain: Norheim et al. \cite{norheim2024challenges} and Hemmat et al. \cite{hemmat2025research} caution that LLMs often struggle with the rigorous logic and domain-specific constraints required for engineering specifications.

The transition to ``Embodied AI'' further complicates this \cite{duan2022survey}. Works like PaLM-E \cite{driess2023palm}, SayCan \cite{brohan2023can}, and Inner Monologue \cite{huang2023inner} demonstrate that LLMs can plan by grounding language into affordances. Yet, a fundamental limitation persists: the Symbol Grounding Problem \cite{bender2020climbing, bisk2020experience}. Lacking embodied experience, models struggle with physical reasoning. Benchmarks like NEWTON \cite{wang2023newton} reveal failures in modeling attributes like mass and friction, while Valmeekam et al. \cite{valmeekam2022large} show deficits in tracking dynamic states. Ruis et al. \cite{ruis2024procedural} further suggest that pre-training data often lacks the procedural knowledge needed for such reasoning. Consequently, reliance on statistical correlations leads to shortcut learning \cite{geirhos2020shortcut}, causing models to overlook long-tail risks \cite{kandpal2023large}. In safety-critical contexts, this disconnect is fatal \cite{kalai2024calibrated}.

\subsection{Automated Hazard Analysis}
Recent research has begun to integrate LLMs into the hazard analysis pipeline. Frameworks like Aegis \cite{shi2024aegis}, Co-Hazard Analysis \cite{diemert2023can}, and prompt-layered approaches \cite{iyenghar2025evaluation} utilize LLMs as ``expert surrogates'' to streamline the generation of safety documentation or map identified hazards to regulations \cite{nouri2024welcome}. While effective for improving the efficiency of traditional safety engineering, these methods remain tethered to the assumption of a finite, pre-defined functional scope. They typically take a static list of robot functions (e.g., ``Navigate'', ``Lift'') as input and retrieve associated risks based on historical data. However, this enumeration-based paradigm fundamentally fails in Embodied AI, where robot behaviors are not pre-scripted but are dynamically instantiated by open-ended user instructions. Existing methods cannot handle the generative variability of natural language commands, where the risk context is constructed on-the-fly. Similarly, while benchmarks like Safety Gymnasium \cite{ji2023safety} provide standardized environments for safe reinforcement learning, they typically operate with pre-defined safety constraints. To bridge the gap between open-ended instructions and physical safety, we propose ARGOS (Attribute-Guided Combinatorial Reasoning). ARGOS shifts the paradigm from implicit ``Label-to-Label'' retrieval to explicit Attribute-Based Deduction, enabling the discovery of complex, multi-factor hazards that purely semantic approaches miss.

\section{Our Framework: ARGOS}
\label{sec:methodology}
We propose an Attribute-Guided Combinatorial Reasoning, a pipeline designed to bridge the gap between abstract user instructions and FSRs. Unlike end-to-end approaches that rely solely on the implicit, hallucination-prone knowledge of LLMs, our framework adopts a structured, physically-anchored two-stage pipeline:

(1) \textbf{Attribute-Guided Hazard Discovery}, which first decomposes semantic entities into granular risk-critical physical properties via Attribute-Centric Contextual Decomposition, and subsequently injects these attributes into a risk-rule context to systematically deduce hazard scenarios through Combinatorial Risk Exploration.

(2) \textbf{Scenario-Anchored Requirement Synthesis}, which employs Constraint-Guided Synthesis to translate these fine-grained scenarios into rigorous engineering specifications aligned with regulatory standards.

This pipeline ensures that safety requirements are derived not from generic linguistic correlations, but from a deductive chain of physical attribution, combinatorial interaction, and regulatory alignment.

The overall architecture of the proposed ARGOS framework is illustrated in Figure \ref{fig:overall_framework}. It visualizes the systematic flow from initial semantic parsing and attribute injection to the final synthesis of physics-aware, testable safety requirements, highlighting the interplay between the two main stages.

\subsection{Stage I:Attribute-Guided Hazard Discovery}
\label{subsec:hazard_discovery}

The objective of this stage is to bridge the gap between high-level user instructions and concrete physical risks. We implement a retrieve-and-infer mechanism to explicitly model the vulnerability of entities and the dynamics of the environment.

\paragraph{Semantic Parsing and Attribute Injection.}
\label{para:parsing_injection}
The process begins with an open-ended user instruction, formalized as a seed scenario $S_{seed}$ (e.g., Deliver hot soup while a child is playing). The system first employs a syntactic parser (spaCy) to extract key semantic units $U = \{u_1, u_2, \dots, u_n\}$. These units serve as the fundamental risk factors (e.g., \texttt{Child, Delivery}).

To eliminate ambiguity, we utilize a BGE-based embedding model to map each unit $u_i$ to its corresponding risk attributes in a pre-defined Rule Base.

The retrieval process is governed by a similarity threshold $\tau_{attr} = 0.7$:
\begin{equation}
A_{injected} = { \text{Attr}(r) \mid r \in \mathcal{R}, \text{sim}(E(u_i), E(r)) > \tau_{attr} }
\end{equation}
The target Rule Base $\mathcal{R}$ is initialized through expert curation to ensure strict adherence to physical laws; representative examples of these rules are provided in Appendix~\ref{app:Rule base}. While this represents a one-time engineering cost, it decouples atomic physical constraints from specific scenarios, allowing the pipeline to automatically synthesize an unbounded number of complex interaction risks without further human intervention.

This step injects domain-specific priors into the context. For instance, the unit \texttt{Child} is explicitly mapped to dynamic attributes like \texttt{High Lateral Acceleration} and \texttt{Unpredictability}, while \texttt{Delivery} implies constraints such as \texttt{Delivery stability} and \texttt{Gripper Compliance}.

\paragraph{Combinatorial Hazard Inference}
\label{subsubsec:risk_imagination}
We construct a Contextual Risk Reasoning task for the LLM. The input consists of the original seed $S_{seed}$ and the injected attribute set $A_{injected}$. The model is prompted to perform a combinatorial deduction to identify how distinct risk factors interact.

    Formally, for a set of $k$ risk factors $\{u_1, \dots, u_k\}$, the hazard mechanism $H_{desc}$ is derived as:
\begin{equation}
H_{desc} = \text{LLM}_{infer} \left( S_{seed}, \bigotimes_{i=1}^{k} A(u_i) \right)
\end{equation}
where $\bigotimes$ denotes a semantic interaction operator that composes individual physical attributes into a joint relational context for the LLM.

As demonstrated in our experiments, for the risk pair \texttt{Child + Delivery}, the model deduces a complex dynamic conflict:
Because the child's high lateral acceleration (Factor A) occurs during a precise delivery maneuver (Factor B), the robot's necessary emergency stop creates inertial forces. These forces overcome the fluid stability maintained by the compliant grip, causing the hot soup to slosh and scald the child.
Crucially, this identifies that the hazard arises not from a simple collision, but from the coupling of the robot's reflex latency and the fluid's inertial dynamics.
\subsection{Stage II: Scenario-Anchored Requirement Synthesis}
\label{subsec:fsr_synthesis}
Stage II converts the descriptive hazard $H_{desc}$ derived in Stage I into prescriptive, testable FSRs. We prioritize the synthesis of FSRs over TSRs to leverage the logical abstraction necessary for LLM reasoning while bypassing the parametric hallucinations inherent in low-level hardware synthesis. Unlike open-ended generation, we formulate this process as a Constraint-Satisfied Synthesis task to enforce alignment with regulatory standards and hardware specifications. Under this framework, FSRs define the governing safety logic; once these functional boundaries are established, the subsequent derivation of specific TSRs is reduced to a deterministic engineering task constrained by the predefined functional bounds.

\paragraph{Regulatory Alignment \& Constraint Injection}
To bridge the gap between abstract risk descriptions and engineering specifications, we first construct a composite constraint context $\mathcal{C}_{ctx}$. This involves retrieving relevant safety standards (e.g., ISO 13482) based on the semantic embedding of the hazard:
\begin{equation}
    \mathcal{R}_{rel} = \{ r \in \mathcal{KB}_{reg} \mid \text{sim}(E(H_{desc}), E(r)) > \tau_{reg} \}
\end{equation}
Simultaneously, we inject the robot's physical specification $\mathcal{K}_{HW}$ (e.g., max deceleration $a_{max}$, sensor range $R_{s}$) into the inference window. This explicitly defines the Feasible Actuation Space for the model.

\paragraph{Physics-Aware Requirement Generation}
The synthesis function $\text{LLM}_{\text{syn}}$ operates on the augmented context to resolve conflicts between safety goals and physical dynamics. The generation of the safety requirement $S_{fsr}$ is modeled as:
\begin{equation}
    S_{fsr} = \text{LLM}_{\text{syn}}\left( H_{desc}, \mathcal{R}_{rel} \mid \mathcal{K}_{HW} \right)
\end{equation}
By focusing on functional logic, the model establishes a logical bridge; once these functional objectives are defined, deriving specific TSRs becomes a straightforward engineering mapping rather than a complex reasoning task. Crucially, the model is prompted to perform a counterfactual feasibility check: it evaluates whether a standard mitigation (e.g., Emergency Stop) violates the physical constraints introduced by $\mathcal{K}_{HW}$.

As observed in the Dynamic Fluid Hazard case, the system successfully identifies a critical conflict: while a collision risk demands braking, the ``Thermal Hazard'' attribute implies that the robot's maximum braking force would trigger fluid instability (sloshing). 
Instead of a generic stop command, the model leverages the deceleration limit $a_{safe} \in \mathcal{K}_{HW}$ to synthesize a sophisticated control policy: 
``When carrying a `Thermal Hazard' and a collision risk exists, the Mobility System shall limit maximum deceleration to $A_{safe}$... prioritizing controlled braking over instantaneous stopping.''
This demonstrates the framework's ability to derive requirements that are not just semantically relevant, but physically grounded.

\section{Experiments}
\subsection{Experimental setup}
\textbf{Scenario Dataset:} A total of 48 ``Seed Scenarios'' involving complex human-robot-environment interactions were curated. These seeds span three critical dimensions: 
(i) Vulnerable Groups (e.g., \texttt{elderly, child, wheelchair users}); 
(ii) Hazardous Environments (e.g., \texttt{low-light, steam-filled rooms}); 
(iii) Risky Objects (e.g., \texttt{sharp tools, hot liquids}). 
Through the proposed inference pipeline, these 48 seeds were expanded into 365 context-specific hazard scenarios.

\textbf{Model Configurations:} \texttt{DeepSeek-V3.2} served as the primary backbone for the complete safety reasoning pipeline. To validate that the reasoning limitations are model-agnostic, we additionally employed \texttt{GPT-4o} as a comparative baseline specifically for the hazard generation stage. \texttt{BGE-Large-EN-v1.5} was used for retrieval, and \texttt{Gemini-3-Pro} functioned as the independent ``LLM-as-a-Judge'' for evaluation, with the evaluation prompts provided in Appendix~\ref{app:prompt_templates}.

\paragraph{Baselines}
To rigorously validate the effectiveness of our proposed pipeline, we compare it against two distinct baseline approaches using the same backbone model (\texttt{DeepSeek-V3.2}).
\begin{itemize}[leftmargin=*, nosep]
\item {Vanilla LLM:} This baseline represents the model's intrinsic safety knowledge. It is prompted to generate hazard scenarios based solely on the seed description and the robot's capability profile, without any intermediate reasoning steps or external attribute grounding. 

\item {Physics-Aware CoT (CoT):}
Unlike generic zero-shot reasoning, this serves as a domain-specific baseline designed to simulate physical dynamics.
We implemented a rigorous prompt (detailed in Appendix~\ref{app:prompt_templates}) that guides the model through a structured three-step reasoning process:
(1) entity decomposition to identify physical vulnerabilities;
(2) micro-action simulation to trace interaction dynamics; and
(3) consequence prediction to evaluate collision risks.
This baseline rigorously tests whether logical reasoning structure alone—without external attribute grounding—is sufficient to uncover long-tail risks.

\item {Attribute-Guided Hazard Discovery (ours):} This pipeline utilizes Granular Risk Injection to interleave domain-specific physical attributes into the generative context. This grounds the inference in high-fidelity scenarios triggered by specific, structured risk factors. We rigorously test the framework's discovery capability across escalating levels of complexity: single-factor, dual-factor, and triple-factor coupled risk interactions.

\end{itemize}
\paragraph{Evaluation Metrics}To rigorously assess the effectiveness of the proposed framework, we define a multi-dimensional evaluation suite categorized into two stages. Stage I focuses on the intrinsic quality and risk depth of the generated hazard scenarios, while Stage II evaluates the system's compliance and logical robustness. A comprehensive summary of these metrics, including their definitions and evaluation objectives, is provided in Table~\ref{tab:metrics_definition}.

\begin{table}[!t]
\centering
\caption{ Definitions of Evaluation Metrics.}
\label{tab:metrics_definition}

\setlength{\aboverulesep}{0pt}
\setlength{\belowrulesep}{0pt}
\renewcommand{\arraystretch}{1.2} 
\footnotesize 
\begin{tabularx}{\columnwidth}{l X}
\toprule
\rowcolor[gray]{0.9} \textbf{Abbr.} & \textbf{Metric Description} \\ \midrule

\multicolumn{2}{l}{\cellcolor{MethodBlueBg!60}\textbf{Hazard Scenario Quality}} \\ 
\textbf{PR} & \textbf{Physical Reliability:} Adherence to real-world physics and strict maintenance of ``closed-world'' constraints. \\
\textbf{LR} & \textbf{Long-tail Risk:} Degree of uncovering statistically rare, concealed, or system-boundary hazards. \\
\textbf{FSR} & \textbf{FSR Derivation:} Actionability in generating specific engineering specifications and safety requirements. \\
\addlinespace[0.3em]

\multicolumn{2}{l}{\cellcolor{MethodBlueBg!60}\textbf{ FSR Quality}} \\ 
\textbf{CC} & \textbf{Capability Compliance:} Respecting physical limits (e.g., blind zones) and leveraging advanced sensors (e.g., thermal/tactile). \\
\textbf{PRC} & \textbf{Physical Risk Coverage:} Exhaustive coverage of primary hazards, secondary consequences, and complex edge cases. \\
\textbf{LRC} & \textbf{Logic Robustness:} Logical closure, clear entry/exit conditions, and state persistence for perception gaps. \\ \bottomrule
\end{tabularx}
\end{table}

\paragraph {Human Evaluation Protocol.}
To evaluate the accuracy of our automated metrics, we invited three human reviewers to conduct a blind review of the generated risk scenarios.

We randomly sampled $N=108$ derived scenarios across three methods (Ours, CoT, Vanilla). 
The humans independently scored each scenario on a scale of 1-10 based on the same three criteria (PR, LR, FSR) used by the LLM judge. 
To mitigate subjective bias, the final human score is reported as the average of the three human reviewers, and we analyze the inter-rater agreement and human-AI alignment in Section~\ref{sec:human_eval}.

\subsection{ Hazard Inference Quality and Diversity}

Table~\ref{tab:stage1_detailed} presents the quantitative results. Our method consistently achieves the highest scores across both backbones. Specifically for Long-tail Risk, the CoT baseline scores lower than the Vanilla baseline on both models. For \texttt{DeepSeek-V3.2}, CoT scores $6.57$ compared to Vanilla's $6.73$; for \texttt{GPT-4o}, CoT scores $5.35$ compared to Vanilla's $6.51$. This decline suggests that standard reasoning prompts may constrain the exploration of low-probability events. In contrast, our method (\texttt{DeepSeek}: $8.21$; \texttt{GPT-4o}: $7.90$) effectively mitigates this issue.

Given that consistent performance trends were observed across both backbones (as shown in Table~\ref{tab:stage1_detailed}), we primarily focus on the \texttt{DeepSeek-V3.2} backbone for the subsequent detailed analysis and visualizations (e.g., distribution plots and t-SNE) to avoid redundancy. Corresponding visualizations for \texttt{GPT-4o} are provided in Appendix ~\ref{app:gpt4o_viz}.
\begin{table}[!tbp]
\centering
\caption{Quantitative results for Stage I. Bold indicates best performance. \colorbox{MethodBlueBg}{Blue rows} represent our method.}
\label{tab:stage1_detailed}

\setlength{\aboverulesep}{0pt}
\setlength{\belowrulesep}{0pt}
\renewcommand{\arraystretch}{1.3}

\resizebox{\columnwidth}{!}{
\begin{tabular}{llcc}
\toprule
\textbf{Metric} & \textbf{Method} & \textbf{DeepSeek-V3.2} & \textbf{GPT-4o} \\ \midrule

\multirow{3}{*}{Physical Reliability} & Vanilla & 7.98 \sd{2.27} & 7.99 \sd{1.78} \\
                                       & CoT     & 8.53 \sd{1.46} & 8.08 \sd{1.38} \\
\rowcolor{MethodBlueBg}                & \textbf{Ours} & \best{8.97} \sd{1.45} & \best{9.06} \sd{1.36} \\ 
\addlinespace[0.5em] 

\multirow{3}{*}{Long-tail Risk}       & Vanilla & 6.73 \sd{1.97} & 6.51 \sd{1.75} \\
                                       & CoT     & 6.57 \sd{2.08} & 5.35 \sd{1.94} \\
\rowcolor{MethodBlueBg}                & \textbf{Ours} & \best{8.21} \sd{1.44} & \best{7.90} \sd{1.81} \\ 
\addlinespace[0.5em]

\multirow{3}{*}{Functional Safety}    & Vanilla & 7.45 \sd{2.08} & 7.22 \sd{1.72} \\
                                       & CoT     & 7.07 \sd{2.03} & 6.03 \sd{1.86} \\
\rowcolor{MethodBlueBg}                & \textbf{Ours} & \best{8.55} \sd{1.47} & \best{8.27} \sd{1.74} \\ 
\bottomrule
\end{tabular}
}
\end{table}

\paragraph{Reasoning vs. Attribute grounded.}
A critical insight from Table~\ref{tab:stage1_detailed} is the performance discrepancy of the CoT baseline. While CoT enhances Physical Reliability (
8.52) compared to the Vanilla LLM (7.98) by enforcing logical structure, it fails to improve and even slightly degrades performance in Long-tail Risk (6.57). This result underscores a fundamental limitation: logical reasoning cannot substitute for domain knowledge.
Without explicit prior knowledge, much like knowing a
rose'' is a flower but neglecting that it has 
thorns''—CoT merely generates generic scenarios. In contrast, our Attribute-Grounded mechanism explicitly retrieves and injects detailed attribute descriptions. This grounds the inference in granular reality, enabling the discovery of non-trivial, domain-specific hazards that pure reasoning overlooks.

\begin{figure}[t] 
    \centering
    \includegraphics[width=1\linewidth]{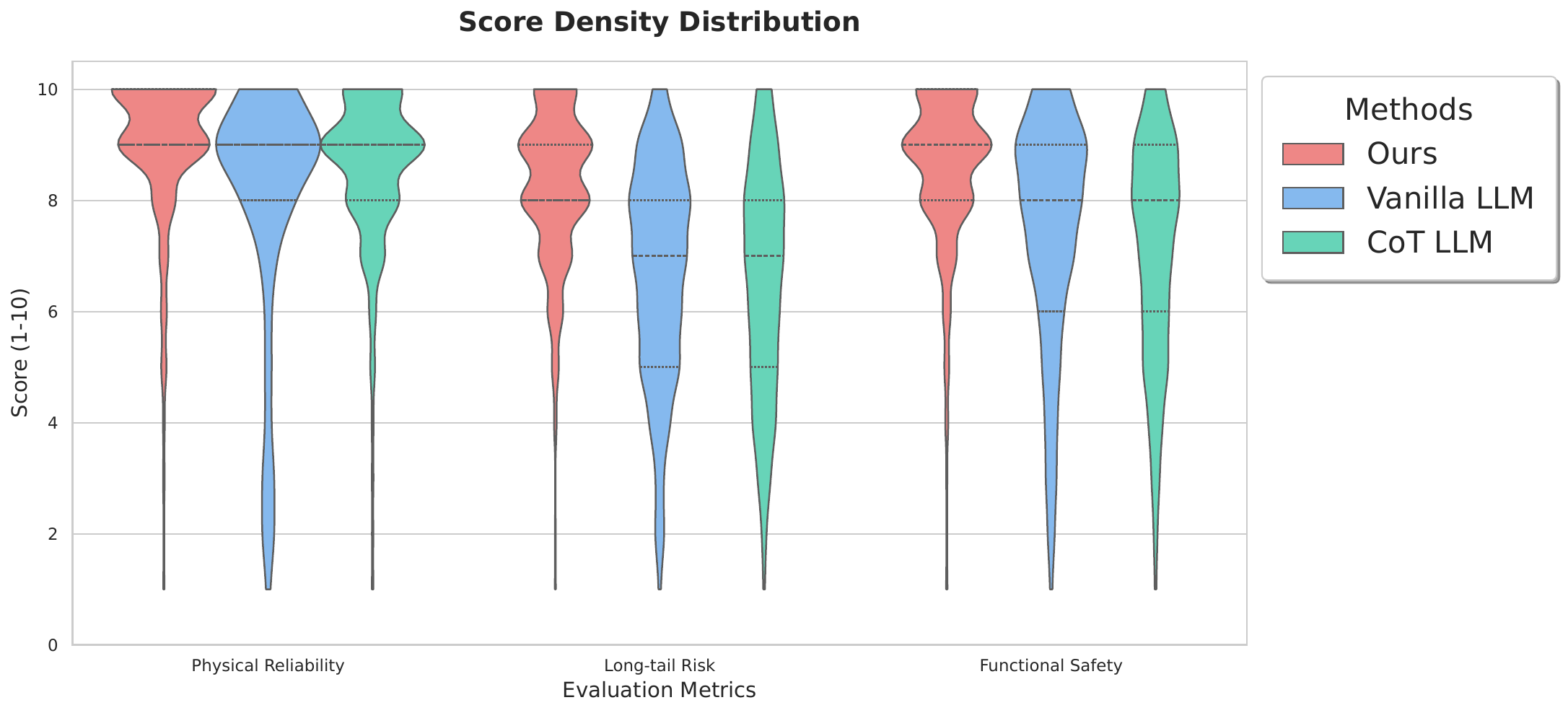}

    \caption{\textbf{Statistical analysis of generation quality.} 
    The violin plot illustrates the score density across methods. 
     }
    \label{fig:violin}
    
\end{figure}

\paragraph{Distributional Robustness.}
Fig.~\ref{fig:violin} corroborates the stability of our pipeline. The violin plot reveals a distinct ``top-heavy'' distribution for our method, with probability mass concentrated in the high-quality (8--10) range. Conversely, baselines exhibit long tails extending into lower scores, reflecting high variance and frequent hallucinations. This confirms that our framework effectively filters out low-fidelity outputs, consistently yielding high-value safety cases.

\paragraph{Latent Semantic Topology Analysis}
\label{sec:semantic_topology}

To rigorously evaluate the geometric properties of the generated risk scenarios, we analyze the latent semantic embeddings using five complementary metrics. 
We report the Effective Rank (Eff. Rank) to estimate the intrinsic dimensionality of the scenario manifold, and Centroid Shift (Shift) to quantify the Euclidean deviation from the seed's Operational Design Domain (ODD). 
Spatial dispersion is measured by Aligned Variance (Aligned Var.). 
Crucially, to distinguish valid exploration from semantic hallucination, we introduce two structural indicators: 
Constrained Semantic Expansion (CSE), defined as the ratio of internal diversity to centroid shift ($\text{Diversity}/\text{Drift}$), which evaluates exploration efficiency; and
Directional Similarity, calculated as the average cosine similarity of difference vectors relative to the seed, where lower values indicate more statistically independent failure modes.

\begin{table}[h] 
\caption{Latent Semantic Topology. While CoT exhibits high raw dimensionality (Rank), its high Directional Similarity and Shift indicate redundant drift. Ours achieves superior CSE and independence (Directional Similarity).}
\label{tab:semantic_metrics}
\centering
\small
\setlength{\aboverulesep}{0pt}
\setlength{\belowrulesep}{0pt}
\renewcommand{\arraystretch}{1.2}
\resizebox{\columnwidth}{!}{
\begin{tabular}{lccccc}
\toprule
{\sc Method} & {\sc Eff. Rank} ($\uparrow$) & {\sc Shift} ($\downarrow$) & {\sc Aligned Var.} ($\uparrow$) & {\sc CSE} ($\uparrow$) & {\sc Directional Similarity} ($\downarrow$) \\
\midrule
Vanilla & 63.66 & 0.000 & 0.111 & N/A$^{\dagger}$ & -0.002 \\
CoT     & \best{67.10} & 0.218 & \best{0.181} & 3.49 & 0.137 \\
\rowcolor{MethodBlueBg} \textbf{Ours} & 66.71 & \best{0.055} & 0.162 &\best{13.14} & \best{0.008} \\
\bottomrule
\multicolumn{6}{l}{\scriptsize $^{\dagger}$Vanilla serves as the anchor (Shift $\approx 0$), rendering the expansion ratio undefined.}
\end{tabular}
}
\end{table}

The quantitative results are presented in Table~\ref{tab:semantic_metrics}. 
A superficial reading of Eff. Rank suggests that CoT(67.10) achieves the highest dimensionality. However, structural metrics reveal that this variance is largely driven by semantic hallucination. 
CoT exhibits a substantial Shift (0.218) combined with high Directional Similarity(0.137). This indicates that CoT scenarios drift away from the seed's ODD in correlated directions. For instance, in the ``Mom fetching disinfectant'' scenario, CoT narratives frequently devolve into generic ``slip and fall'' accidents—semantically distant from the task (High Shift) yet highly redundant (High Directional Similarity).

In contrast, Ours maintains a tight semantic proximity to the seed (Shift 0.055) while achieving a Rank (66.71) comparable to CoT. 
Crucially, our method demonstrates superior structural properties: near-zero Directional Similarity (0.008) confirms that the generated failure modes (e.g., LiDAR blind zones vs. chemical corrosion) are statistically independent. Consequently, Ours achieves a CSE score of 13.14, which is \textbf{3.76} higher than CoT (3.49). 
This confirms that our method performs constrained exploration: maximizing the discovery of distinct, valid risk scenarios without suffering from the semantic drift observed in chain-of-thought baselines.

\begin{figure*}[t]
    \centering
    
    \begin{subfigure}[b]{0.39\textwidth}
        \centering
        \includegraphics[width=\linewidth]{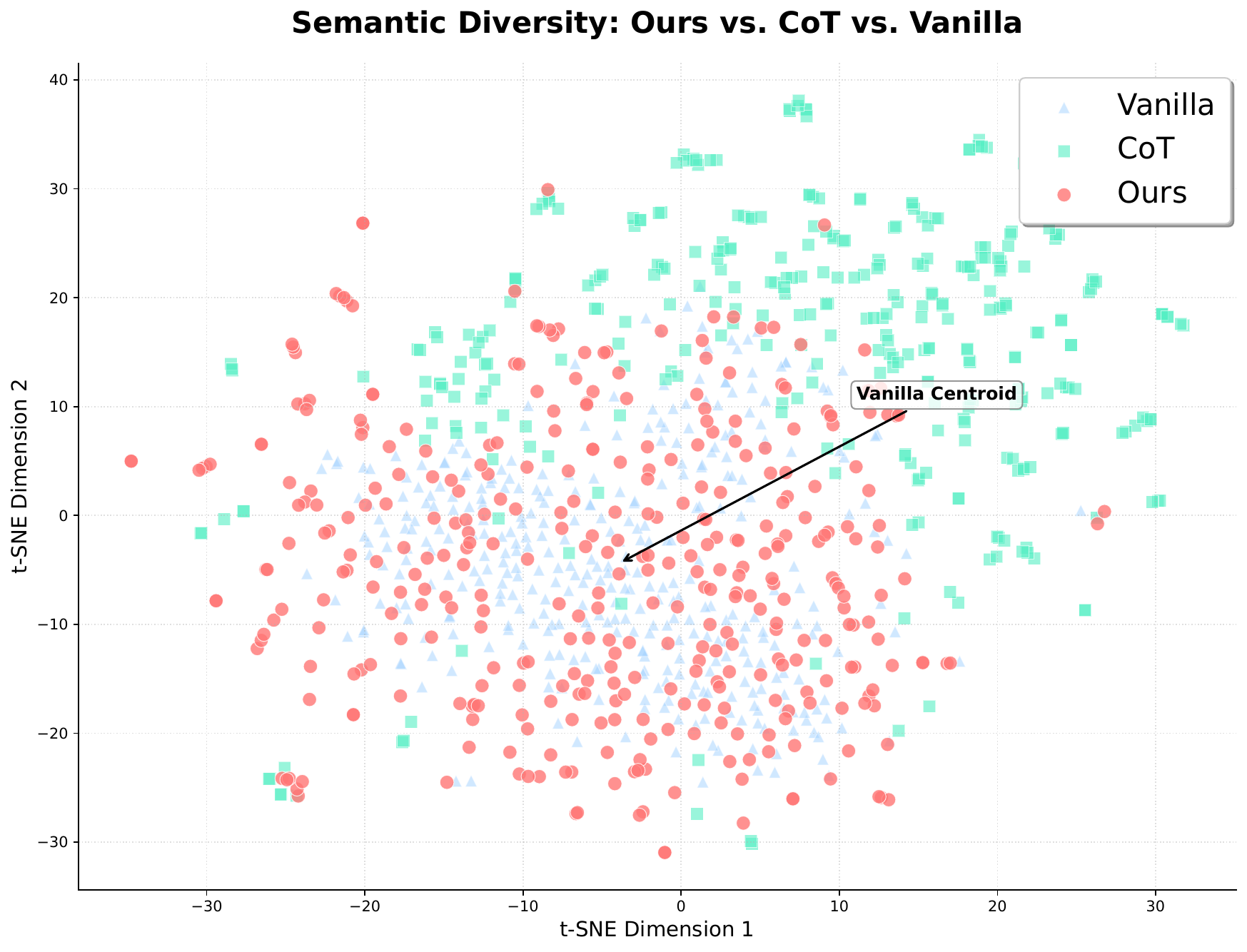}
        \caption{\textbf{t-SNE visualization of scenario embeddings}}
        \label{fig:tsne} 
    \end{subfigure}
    \hfill 
    \begin{subfigure}[b]{0.59\textwidth}
        \centering
        \includegraphics[width=\linewidth]{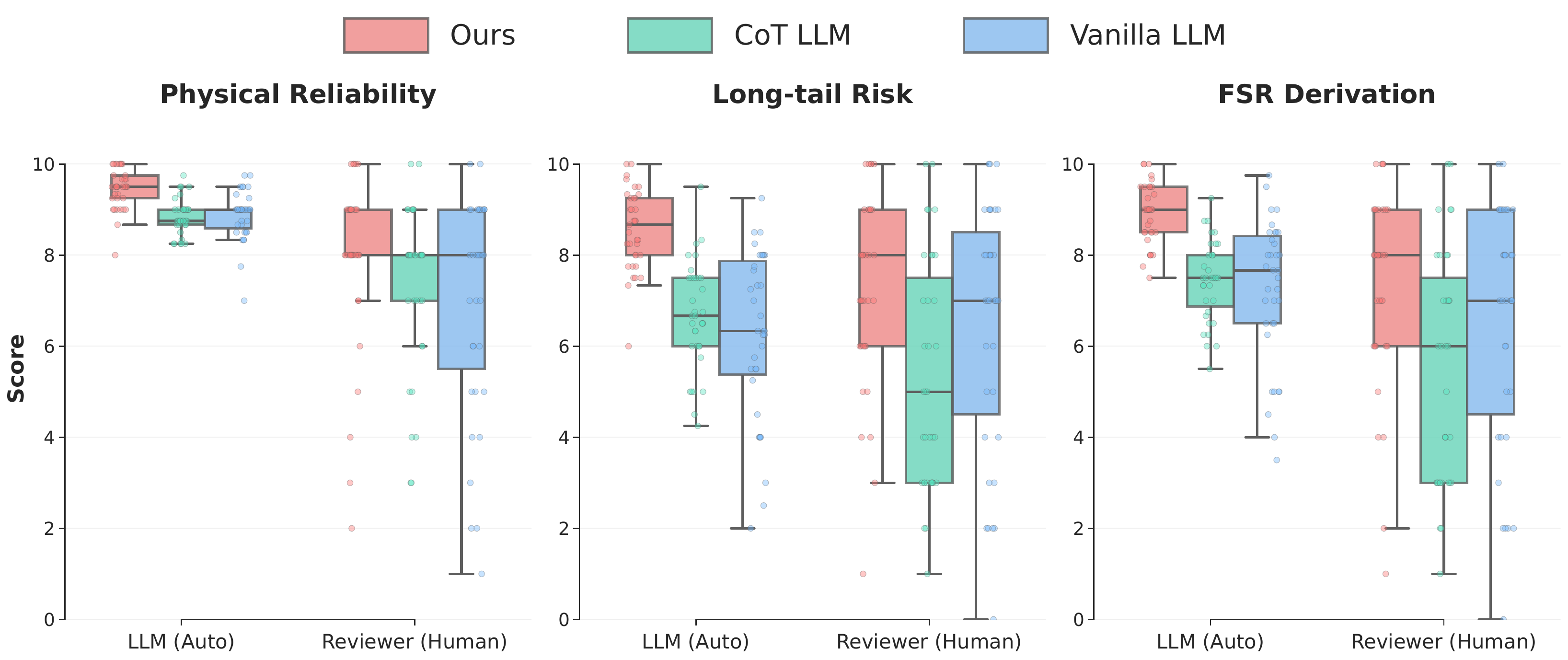} 
        \caption{\textbf{Human-AI Score Distribution Alignment.} While LLMs exhibit a systematic ``positivity bias'' (inflated absolute scores), the relative performance ranking ($\text{Ours} > \text{Vanilla} \approx \text{CoT}$) remains consistent. Crucially, humans penalized CoT heavily for semantic redundancy, a flaw overlooked by the LLM judge.}
        \label{fig:expert_boxplot} 
    \end{subfigure}
    
    \caption{Qualitative Analysis: Semantic Diversity and Evaluation Alignment.}
    \label{fig:qualitative_analysis} 
\end{figure*}

\paragraph{Human-AI Alignment and human Validation}
\label{sec:human_eval}

To validate the reliability of the automated ``LLM-as-a-Judge'' metrics, we conducted a rigorous human study with reviewers having technical backgrounds in safety-critical systems. We randomly sampled 108 derived sub-scenarios and invited three human reviewers to conduct a blind evaluation.

The comparative results are summarized in Table~\ref{tab:human_expert}. Analyzing the discrepancy between LLM and human scores reveals two critical scientific insights:
\textbf{1. Systematic Calibration Offset vs. Relative Consistency:} 
As shown in Fig.~\ref{fig:expert_boxplot}, we observe a significant calibration offset: LLM judges consistently overestimate scenario quality by $\sim$1.5 points. This aligns with findings on the ``generosity bias'' of LLM evaluators. While the LLM judge reliably identifies our method as the superior one with a clear margin, its performance on baselines (Vanilla vs. CoT) is less decisive. The scores for these two baselines are clustered closely in both evaluation regimes, leading to minor fluctuations in their relative ranking.

\textbf{2. The ``Verbosity Trap'' in CoT Reasoning:} 
A noteworthy divergence occurs in the Long-tail Risk (LR) metric. Although the scores for Vanilla and CoT are numerically close, the LLM judge showed a slight preference for CoT (6.66 vs 6.24), whereas humans leaned toward Vanilla (6.43 vs 5.29). Qualitative analysis of the human logs reveals that CoT suffers from a ``Verbosity Trap'': it generates logically structured but semantically repetitive narratives (labeled by humans as ``Correct Nonsense''). For instance, in the Hot Water Delivery case, CoT repeated variations of ``slipping risk'' six times across different steps, lacking the causal depth to identify system-level conflicts. In contrast, our method achieves the highest human LR score ($7.26 \pm 2.12$) by discovering complex coupled risks, such as the ``Safety Pinch'' paradox (where a safety stop triggers a secondary fluid hazard).

\begin{table}[h]
\centering
\caption{Comparison of Automated (LLM) vs. Human  Evaluation. Note the significant drop in CoT's humans scores for Long-tail Risk. humans penalized CoT for generating generic, repetitive scenarios (``Correct Nonsense'') that fooled the LLM judge.}
\label{tab:human_expert}
\setlength{\aboverulesep}{0pt}
\setlength{\belowrulesep}{0pt}
\renewcommand{\arraystretch}{1.2}
\resizebox{\columnwidth}{!}{
\begin{tabular}{l|cc|cc|cc}
\toprule
\multirow{2}{*}{\textbf{Method}} & \multicolumn{2}{c|}{\textbf{Physical Reliability (PR)}} & \multicolumn{2}{c|}{\textbf{Long-tail Risk (LR)}} & \multicolumn{2}{c}{\textbf{FSR Derivation}} \\
 & LLM & Human & LLM & Human & LLM & Human \\ 
\midrule
Vanilla LLM & 8.88 \scriptsize{$\pm 0.53$} & 6.97 \scriptsize{$\pm 2.40$} & 6.24 \scriptsize{$\pm 1.82$} & 6.43 \scriptsize{$\pm 2.70$} & 7.21 \scriptsize{$\pm 1.57$} & 6.49 \scriptsize{$\pm 2.67$} \\
CoT LLM     & 8.84 \scriptsize{$\pm 0.38$} & 7.34 \scriptsize{$\pm 1.79$} & 6.66 \scriptsize{$\pm 1.16$} & 5.29 \scriptsize{$\pm 2.51$} & 7.45 \scriptsize{$\pm 0.87$} & 5.46 \scriptsize{$\pm 2.53$} \\
\rowcolor{MethodBlueBg} \textbf{Ours} & \best{9.46} \scriptsize{$\pm 0.44$} & \best{7.94} \scriptsize{$\pm 1.87$} & \best{8.56} \scriptsize{$\pm 0.87$} & \best{7.26} \scriptsize{$\pm 2.12$} & \best{8.91} \scriptsize{$\pm 0.65$} & \best{7.34} \scriptsize{$\pm 2.12$} \\ 
\bottomrule
\end{tabular}
}
\end{table}

\subsection{ Evaluation of FSR Quality}
\label{subsubsec:stage2_eval}

Table~\ref{tab:stage2_results} evaluates the contribution of ARGOS components through an ablation study. Comparing our complete framework against w/o ISO, ISO-Only, and Vanilla variants, we observe that explicit attribute-driven deduction significantly outperforms implicit statistical retrieval. The results (Mean 
± SD) demonstrate that ARGOS enhances both generation quality and stochastic stability.

\begin{table}[t]
\centering
\caption{Ablation Study on FSRs Quality. ARGOS (Ours) denotes the full pipeline. w/o ISO removes regulatory retrieval.ISO-Only removes attribute injection. Vanilla is the baseline. Metrics are reported as Mean $\pm$ SD.}
\label{tab:stage2_results}

\setlength{\aboverulesep}{0pt}
\setlength{\belowrulesep}{0pt}
\renewcommand{\arraystretch}{1.3}

\resizebox{\columnwidth}{!}{
\begin{tabular}{lcccc}
\toprule
\textbf{Method} & \textbf{CC} $\uparrow$ & \textbf{PRC} $\uparrow$ & \textbf{LRC} $\uparrow$ & \textbf{Overall} $\uparrow$ \\ \midrule
Vanilla       & 7.91 \sd{1.95} & 6.96 \sd{1.92} & 7.28 \sd{2.02} & 7.38 \sd{1.66} \\
ISO-Only      & 7.96 \sd{2.19} & 7.00 \sd{2.03} & 6.90 \sd{1.95} & 7.28 \sd{1.68} \\
w/o ISO       & 8.56 \sd{1.97} & 7.71 \sd{1.70} & 7.29 \sd{2.27} & 7.85 \sd{1.70} \\

\addlinespace[0.5em] 
\rowcolor{MethodBlueBg} 
\textbf{ARGOS (Ours)} & \best{8.56} \sd{1.47} & \best{7.81} \sd{1.71} & \best{7.54} \sd{1.70} & \best{7.97} \sd{1.25} \\ 
\bottomrule
\end{tabular}
}
\end{table}

\noindent \textbf{Physical Grounding as the Precondition for Compliance.} 
The results for Hardware Compliance provide empirical evidence for the Symbol Grounding Problem discussed in Section 1. Notably, w/o ISO and ARGOS achieve an identical mean score of 8.56, significantly outperforming ISO-Only (7.96). This confirms that the capability to identify hardware constraints (e.g. braking distance limits) is strictly conditioned on the injection of physical attributes rather than the retrieval of regulatory text. Without the shared experience of physical laws provided by our attribute decomposition, the ISO-Only model fails to anchor abstract clauses to the scenario, resulting in high variance (SD=2.19). In contrast, the inclusion of ISO standards in ARGOS acts as a regularizer, reducing the SD to 1.47 by constraining the output format, thus ensuring engineering consistency.

\noindent \textbf{Overcoming Shortcut Learning in Risk Coverage.} 
The superior performance of ARGOS in Physical Risk Coverage (7.81) validates the effectiveness of our Combinatorial Hazard Simulation. While the Vanilla baseline (6.96) relies on ``Label-to-Label'' shortcut learning, identifying only generic risks like collisions, our method leverages attribute-based deduction to uncover long-tail hazards.

A compelling example is the ``Rose \& Balloon'' scenario. The baseline, treating the rose as a generic obstacle, merely prescribes a kinematic constraint (e.g., ``Limit velocity $<$ 0.1 m/s''), which fails to prevent the rose's thorns from popping the fragile balloon even at low speeds. In contrast, ARGOS explicitly retrieves the ``Thorn'' attribute, reasoning that contact risk exists regardless of velocity. Consequently, it synthesizes a dynamic constraint (``Limit end-effector force''), prioritizing interaction safety over simple collision avoidance.

This confirms that explicit attribute injection provides a robust cognitive scaffolding, preventing the model from overlooking subtle, multi-factor risks.

\noindent \textbf{The "Regulatory Mismatch" Phenomenon.} 
A critical finding in Logic Robustness is that ISO-Only (6.90) underperforms the Vanilla baseline (7.28). We term this the Regulatory Mismatch phenomenon. When rigorous safety standards are applied to a vague, semantic-only context, the model suffers from a grounding failure, hallucinating conflicting trigger conditions to satisfy the regulatory text. In contrast, ARGOS achieves the highest score (7.54) with the lowest variance (1.70). This demonstrates that regulatory retrieval is only effective when paired with high-fidelity physical context. The physical attributes provide the necessary variables for the ISO standards to latch onto, transforming abstract rules into actionable specifications.
\section{Conclusion and Discussion}
\label{sec:conclusion}
We present ARGOS, the first automated pipeline for synthesizing FSRs from open-ended instructions. To address HARA’s scalability limits and LLM grounding failures, ARGOS shifts from implicit mapping to explicit Attribute-Guided Combinatorial Reasoning. By decomposing entities into risk factors and anchoring them to regulatory standards, it bridges the gap between semantic commands and physical constraints. This causal approach effectively uncovers long-tail risks and outperforms baselines, offering a scalable path for verifiable Embodied AI safety.

Beyond technical efficacy, we observe that rigid adherence to current standards (e.g., ISO 13482:2014) may occasionally constrain reasoning, suggesting a need for evolving safety protocols. Furthermore, ARGOS is envisioned as a "Human-in-the-Loop" assistant to support expert decision-making rather than a standalone replacement. Future work will integrate Vision-Language Models (VLMs) to ground safety reasoning in dynamic, real-world visual perceptions.




\bibliography{references}

@inproceedings{shi2024aegis,
    title     = {{Aegis}: An Advanced {LLM}-Based {Multi-Agent} for Intelligent {Functional Safety} Engineering},
    author    = {Shi, Lu and Qi, Bin and Luo, Jiarui and Zhang, Yang and Liang, Zhanzhao and Gao, Zhaowei and Deng, Wenke and Sun, Lin},
    booktitle = {Proceedings of the 2024 Conference on Empirical Methods in Natural Language Processing (EMNLP): Industry Track},
    year      = {2024},
    pages     = {1571--1583}
}

@article{2017LasotaSurvey,
  title={A survey of methods for safe human-robot interaction},
  author={Lasota, Przemyslaw A and Fong, Terrence and Shah, Julie A and others},
  journal={Foundations and Trends{\textregistered} in Robotics},
  volume={5},
  number={4},
  pages={261--349},
  year={2017},
  publisher={Now Publishers, Inc.}
}

@inproceedings{driess2023palm,
  title={{PaLM-E: An Embodied Multimodal Language Model}},
  author={Driess, Danny and Xia, Fei and Sajjadi, Mehdi SM and Lynch, Corey and Chowdhery, Aakanksha and Ichter, Brian and Wahid, Ayzaan and Tompson, Jonathan and Vuong, Quan and Yu, Tianhe and others},
  booktitle={Proceedings of the 40th International Conference on Machine Learning (ICML)},
  pages={8469--8488},
  year={2023}
}

@inproceedings{brohan2023can,
  title={Do as {I} can, not as {I} say: Grounding language in robotic affordances},
  author={Brohan, Anthony and Chebotar, Yevgen and Finn, Chelsea and Hausman, Karol and Herzog, Alexander and Ho, Daniel and Ibarz, Julian and Irpan, Alex and Jang, Eric and Julian, Ryan and others},
  booktitle={Conference on Robot Learning (CoRL)},
  pages={287--318},
  year={2023}
}

@article{seshia2022toward,
  title={Toward verified artificial intelligence},
  author={Seshia, Sanjit A and Sadigh, Dorsa and Sastry, S Shankar},
  journal={Communications of the ACM},
  volume={65},
  number={7},
  pages={46--55},
  year={2022},
  publisher={ACM New York, NY, USA}
}

@inproceedings{kalai2024calibrated,
  title={Calibrated language models must hallucinate},
  author={Kalai, Adam Tauman and Vempala, Santosh S},
  booktitle={Proceedings of the 56th Annual ACM Symposium on Theory of Computing (STOC)},
  pages={160--171},
  year={2024}
}

@inproceedings{ruis2024procedural,
  title={Procedural Knowledge in Pretraining Drives Reasoning in Large Language Models},
  author={Ruis, Laura and Mozes, Maximilian and Bae, Juhan and Kamalakara, Siddhartha Rao and Gnaneshwar, Dwaraknath and Locatelli, Acyr and Kirk, Robert and Rockt{\"a}schel, Tim and Grefenstette, Edward and Bartolo, Max},
  booktitle={The Thirteenth International Conference on Learning Representations (ICLR)},
  year={2024}
}

@article{hemmat2025research,
  title={Research directions for using {LLM} in software requirement engineering: A systematic review},
  author={Hemmat, Arshia and Sharbaf, Mohammadreza and Kolahdouz-Rahimi, Shekoufeh and Lano, Kevin and Tehrani, Sobhan Y},
  journal={Frontiers in Computer Science},
  volume={7},
  pages={1519437},
  year={2025},
  publisher={Frontiers Media SA}
}

@inproceedings{nouri2024welcome,
  title={Welcome your new {AI} teammate: On safety analysis by leashing large language models},
  author={Nouri, Ali and Cabrero-Daniel, Beatriz and Torner, Fredrik and Sivencrona, Hakan and Berger, Christian},
  booktitle={Proceedings of the 3rd IEEE/ACM International Conference on AI Engineering – Software Engineering for AI (CAIN)},
  pages={172--177},
  year={2024}
}

@inproceedings{valmeekam2022large,
  title={Large Language Models Still Can't Plan ({A} Benchmark for {LLMs} on Planning and Reasoning about Change)},
  author={Valmeekam, Karthik and Olmo, Alberto and Sreedharan, Sarath and Kambhampati, Subbarao},
  booktitle={NeurIPS 2022 Workshop on Foundation Models for Decision Making},
  year={2022}
}

@inproceedings{wang2023newton,
  title={NEWTON: Are Large Language Models Capable of Physical Reasoning?},
  author={Wang, Yi Ru and Duan, Jiafei and Fox, Dieter and Srinivasa, Siddhartha},
  booktitle={The 2023 Conference on Empirical Methods in Natural Language Processing (EMNLP)},
  year={2023}
}

@inproceedings{bisk2020experience,
  title={Experience Grounds Language},
  author={Bisk, Yonatan and Holtzman, Ari and Thomason, Jesse and Andreas, Jacob and Bengio, Yoshua and Chai, Joyce and Lapata, Mirella and Lazaridou, Angeliki and May, Jonathan and Nisnevich, Aleksandr and others},
  booktitle={Proceedings of the 2020 Conference on Empirical Methods in Natural Language Processing (EMNLP)},
  pages={8718--8735},
  year={2020}
}

@inproceedings{bender2020climbing,
title={Climbing towards {NLU}: On meaning, form, and understanding in the age of data},
author={Bender, Emily M and Koller, Alexander},
booktitle={Proceedings of the 58th Annual Meeting of the Association for Computational Linguistics (ACL)},
pages={5185--5198},
year={2020}
}

@inproceedings{kandpal2023large,
  title={Large language models struggle to learn long-tail knowledge},
  author={Kandpal, Nikhil and Deng, Haikang and Roberts, Adam and Wallace, Eric and Raffel, Colin},
  booktitle={Proceedings of the 40th International Conference on Machine Learning (ICML)},
  pages={15696--15707},
  year={2023}
}

@article{geirhos2020shortcut,
  title={Shortcut learning in deep neural networks},
  author={Geirhos, Robert and Jacobsen, J{\"o}rn-Henrik and Michaelis, Claudio and Zemel, Richard and Brendel, Wieland and Bethge, Matthias and Wichmann, Felix A},
  journal={Nature Machine Intelligence},
  volume={2},
  number={11},
  pages={665--673},
  year={2020},
  publisher={Nature Publishing Group UK London}
}

@techreport{iso13482,
  author      = {{International Organization for Standardization}},
  title       = {Robots and Robotic Devices --- Safety Requirements for Personal Care Robots},
  type        = {Standard},
  number      = {ISO 13482:2014},
  year        = {2014},
  address     = {Geneva, Switzerland},
  institution = {International Organization for Standardization}
}

@book{leveson2016engineering,
  title={Engineering a safer world: Systems thinking applied to safety},
  author={Leveson, Nancy G},
  year={2016},
  publisher={The MIT Press}
}

@article{hou2024large,
  title={Large language models for software engineering: A systematic literature review},
  author={Hou, Xinyi and Zhao, Yanjie and Liu, Yue and Yang, Zhou and Wang, Kailong and Li, Li and Luo, Xiapu and Lo, David and Grundy, John and Wang, Haoyu},
  journal={ACM Transactions on Software Engineering and Methodology},
  volume={33},
  number={8},
  pages={1--79},
  year={2024},
  publisher={ACM New York, NY}
}

@article{norheim2024challenges,
  title={Challenges in applying large language models to requirements engineering tasks},
  author={Norheim, Johannes J and Rebentisch, Eric and Xiao, Dekai and Draeger, Lorenz and Kerbrat, Alain and de Weck, Olivier L},
  journal={Design Science},
  volume={10},
  pages={e16},
  year={2024},
  publisher={Cambridge University Press}
}

@inproceedings{fan2023large,
  title={Large language models for software engineering: Survey and open problems},
  author={Fan, Angela and Gokkaya, Beliz and Harman, Mark and Lyubarskiy, Mitya and Sengupta, Shubho and Yoo, Shin and Zhang, Jie M},
  booktitle={2023 IEEE/ACM International Conference on Software Engineering: Future of Software Engineering (ICSE-FoSE)},
  pages={31--53},
  year={2023},
  organization={IEEE}
}

@article{marques2024using,
  title  = {Using {ChatGPT} in Software Requirements Engineering: A Comprehensive Review},
  author={Marques, Nuno and Silva, Rodrigo Rocha and Bernardino, Jorge},
  journal={Future Internet},
  volume={16},
  number={6},
  pages={180},
  year={2024},
  publisher={MDPI}
}

@inproceedings{huang2023inner,
  title={Inner Monologue: Embodied Reasoning through Planning with Language Models},
  author={Huang, Wenlong and Xia, Fei and Xiao, Ted and Chan, Harris and Liang, Jacky and Florence, Pete and Zeng, Andy and Tompson, Jonathan and Mordatch, Igor and Chebotar, Yevgen and others},
  booktitle={Conference on Robot Learning},
  pages={1769--1782},
  year={2023}
}

@inproceedings{ji2023safety,
  title={Safety gymnasium: A unified safe reinforcement learning benchmark},
  author={Ji, Jiaming and Zhang, Borong and Zhou, Jiayi and Pan, Xuehai and Huang, Weidong and Sun, Ruiyang and Geng, Yiran and Zhong, Yifan and Dai, Josef and Yang, Yaodong},
  booktitle={Advances in Neural Information Processing Systems (NeurIPS)},
  volume={36},
  pages={18964--18993},
  year={2023}
}

@article{duan2022survey,
  title={A survey of embodied ai: From simulators to research tasks},
  author={Duan, Jiafei and Yu, Samson and Tan, Hui Li and Zhu, Hongyuan and Tan, Cheston},
  journal={IEEE Transactions on Emerging Topics in Computational Intelligence},
  volume={6},
  number={2},
  pages={230--244},
  year={2022},
  publisher={IEEE}
}

@inproceedings{diemert2023can,
  title={Can large language models assist in hazard analysis?},
  author={Diemert, Simon and Weber, Jens H},
  booktitle={International Conference on Computer Safety, Reliability, and Security (SAFECOMP)},
  pages={410--422},
  year={2023},
  organization={Springer}
}

@article{iyenghar2025evaluation,
  title={Evaluation of Automated Machinery Functional Safety Risk Assessment Using {LLMs}},
  author={Iyenghar, Padma and Mansour, Zaher and Wuebbelmann, Juergen},
  journal={IEEE Access},
  volume={13},
  pages={203648--203669},
  year={2025},
  publisher={IEEE}
}
\bibliographystyle{icml2026}
\newpage
\appendix
\onecolumn


\section{Prompt Construction \& Templates}
\label{app:prompt_templates}

In this section, we detail the exact prompt templates used to instantiate our \textit{Attribute-Grounded Generation} framework. Unlike standard retrieval approaches that treat retrieved text as passive context, our method formulates the generation task as a \textbf{Dual-Factor Constraint Satisfaction Problem}. 

The prompt consists of three modular blocks:
\begin{enumerate}
    \item \textbf{System Definition:} Establishes the robot's physical embodiment (see Appendix~\ref{app:robot_specs}).
    \item \textbf{Risk Injection Block (Ours):} Dynamically embeds two orthogonal risk factors as coupled constraints.
    \item \textbf{Task Specification:} The seed scenario and output formatting rules.
\end{enumerate}

\subsection{Risk Injection Mechanism (Ours)}

\label{app:risk_injection}

The core novelty of our method lies in the \textit{Risk Injection Block}. We retrieve two complementary risk factors ($R_A, R_B$) and inject them into the following template. This forces the LLM to reason about the \textit{interaction} between environmental hazards and robot limitations.

\begin{tcolorbox}[colback=green!5, colframe=green!40, title=\textbf{Template: Dual-Factor Injection Block}]
\begin{lstlisting}[basicstyle=\scriptsize\ttfamily, breaklines=true, columns=fullflexible, xleftmargin=0pt, extendedchars=true]
Template: Dual-Factor Risk Scenario Prompt

[SYSTEM ROLE]
You are a safety engineer identifying how SPECIFIC, REAL-WORLD risk conditions can lead to harm during home robot tasks.

Generate a scenario where a hazard occurs due to the interaction of the specified risk factors.

{Robot_Specifications}

[RISK INTERACTION]
The hazard must be caused by the INTERACTION of TWO specific risk factors.
Condition: The hazard occurs because BOTH factors are present simultaneously.

Risk Factor A: {{Risk_A.Value}} ({{Risk_A.Dimension}})
Definition: "{{Risk_A.Description}}"

Risk Factor B: {{Risk_B.Value}} ({{Risk_B.Dimension}})
Definition: "{{Risk_B.Description}}"

Seed scenario: "{Seed_Scenario}"

[CONSTRAINTS FOR SCENARIO GENERATION]

1.The scenario must remain strictly within the semantic boundary defined by the seed scenario.
2.Only entities explicitly mentioned in the seed or logically indispensable for the task may be used.
3.No unmentioned object, state, event, or environmental condition may be introduced under any circumstance.
4.The root cause must be the coupling effect of Factor A and Factor B.
5.If one factor is removed, the accident must become impossible.
6.Robots will not have emotions.
7.[STRICT CLOSED WORLD] ROBOT CAPABILITIES: Strictly control the robot's behavior within the defined capabilities.
8.STRICT ADHERENCE TO SPECIFICATIONS: Strictly enforce the union of all sensor coverage ranges defined in the robot capabilities.
9.[CRITICAL] MAINTAIN PHYSICAL REALISM.
10.[LOGICAL CONSISTENCY] CAUSAL VALIDITY (DETECTION != PREVENTION).
[OUTPUT FORMAT]
EXACTLY TWO LINES:
Line 1: 2-3 sentences describing the robot's task execution and how the two factors combined to cause the hazard.
Line 2: "Hazard mechanism: [one sentence explaining the coupling effect]."
\end{lstlisting}
\end{tcolorbox}

\subsection{Full System Prompts}

\paragraph{Baseline Prompt (Vanilla LLM)}
The baseline prompt provides the robot capabilities but lacks the specific risk injection block.

\begin{tcolorbox}[colback=gray!10, colframe=gray!50, title=\textbf{System Prompt: Baseline}]
\begin{lstlisting}[basicstyle=\scriptsize\ttfamily, breaklines=true, columns=fullflexible, xleftmargin=0pt, extendedchars=true]
Template: Baseline Hazardous Scenario Generation (CoT)

[SYSTEM ROLE]
You are a safety engineer identifying how real-world conditions can lead to harm during home robot tasks.

Generate a scenario where a hazard occurs due to potential risks inherent in the task or environment.

[ROBOT CAPABILITIES]
{Robot_Specifications}

[SEED SCENARIO]
"{Seed_Scenario}"

[CONSTRAINTS]

1.The scenario must remain strictly within the semantic boundary defined by the seed scenario.
2.Only entities explicitly mentioned in the seed or logically indispensable for the task may be used.
3.Robots will not have emotions similar to humans, such as fear or anxiety, startle.
4.No unmentioned object, state, event, or environmental condition may be introduced under any circumstance.
5.STRICT ADHERENCE TO SPECIFICATIONS: Strictly enforce the union of all sensor coverage ranges defined in the robot capabilities.
6.[STRICT CLOSED WORLD] ROBOT CAPABILITIES: Strictly control the robot's behavior within the defined capabilities.
7.[CRITICAL] MAINTAIN PHYSICAL REALISM.
8.[LOGICAL CONSISTENCY] CAUSAL VALIDITY (DETECTION != PREVENTION).

[OUTPUT FORMAT]
Line 1: 2-3 sentences describing the robot's task execution and how the two factors combined to cause the hazard.
Line 2: "Hazard mechanism: [one sentence explaining the coupling effect].
\end{lstlisting}
\end{tcolorbox}

\subsubsection{Chain-of-Thought (CoT) Prompt}
The CoT baseline appends reasoning instructions to the standard prompt to encourage step-by-step simulation.

\begin{tcolorbox}[colback=blue!5, colframe=blue!40, title=\textbf{Add-on: CoT Reasoning Instructions}]
\begin{lstlisting}[basicstyle=\scriptsize\ttfamily, breaklines=true, columns=fullflexible, xleftmargin=0pt, extendedchars=true]
Template: Baseline Hazardous Scenario Generation (CoT)

[SYSTEM ROLE]
You are a safety engineer identifying how real-world conditions can lead to harm during home robot tasks.

Generate a scenario where a hazard occurs due to potential risks inherent in the task or environment.

[ROBOT CAPABILITIES]
{Robot_Specifications}

[SEED SCENARIO]
"{Seed_Scenario}"

[CONSTRAINTS]

1.The scenario must remain strictly within the semantic boundary defined by the seed scenario.
2.Only entities explicitly mentioned in the seed or logically indispensable for the task may be used.
3.Robots will not have emotions similar to humans, such as fear or anxiety, startle.
4.No unmentioned object, state, event, or environmental condition may be introduced under any circumstance.
5.STRICT ADHERENCE TO SPECIFICATIONS: Strictly enforce the union of all sensor coverage ranges defined in the robot capabilities.
6.[STRICT CLOSED WORLD] ROBOT CAPABILITIES: Strictly control the robot's behavior within the defined capabilities.
7.[CRITICAL] MAINTAIN PHYSICAL REALISM.
8.[LOGICAL CONSISTENCY] CAUSAL VALIDITY (DETECTION != PREVENTION).
[REASONING INSTRUCTIONS]
Before generating the final result, you must analyze the seed scenario step-by-step inside an [ANALYSIS] block:

Entity & Vulnerability Analysis (Decomposition):

Extract all key Entities (Who/What), Actions (Task), and Environmental Context.
Vulnerability Inference: Analyze inherent physical limitations or vulnerabilities.
Micro-Action Simulation (Step-by-Step Trace):

Break down the robot's task into a sequence of logical micro-actions.
Simulate the physical interaction between the robot and identified entities.
Consequence Analysis:

Identify identifies accident chains.
[OUTPUT FORMAT]
[ANALYSIS]

Decomposition: [Entities, Context, and vulnerabilities...]
Simulation: [Step-by-step trace...]
Prediction: [Logical deduction...]
[OUTPUT FORMAT]
Line 1: 2-3 sentences describing the robot's task execution and how the two factors combined to cause the hazard.
Line 2: "Hazard mechanism: [one sentence explaining the coupling effect].
\end{lstlisting}
\end{tcolorbox}

\newpage
\section{Rule base}
\label{app:Rule base}

This appendix provides the formal technical definitions for the dimensions used in the \textit{Risk Injection Block} (see Appendix~\ref{app:risk_injection}). These definitions serve as the ground-truth constraints for the LLM to reason about physical vulnerabilities and environmental hazards.

\subsection{User State Dimensions (U)}
\label{app:user_states}

The following definitions specify the physiological and behavioral constraints of human and non-human agents in the environment.

\begin{tcolorbox}[colback=blue!5, colframe=blue!40, title=\textbf{Table B.1: User State Definitions}]
\footnotesize
\begin{description}[leftmargin=0.5cm, itemsep=2pt]
    \item[\texttt{U01} Elderly:] Human individual aged 65 years or older, exhibiting reduced visual acuity in low illumination, slower neuromuscular response latency to visual motion stimuli, and decreased postural stability during upper-body perturbations.
    \item[\texttt{U13} Toddler:] Human individual (aged 1--3 years) with standing height between 70 cm and 95 cm. Characterized by an unstable gait with high fall frequency, limited hazard recognition, and unpredictable reach-and-grasp motions. Fingers are small enough to enter gaps as narrow as 5 mm. High propensity for ``oral exploration.''
    \item[\texttt{U12} Pet Animal:] Non-human biological agent (e.g., \textit{Canis lupus familiaris}, \textit{Felis catus}) with highly variable morphology, typically ranging from 10--80~cm in height. Characterized by high-DoF omni-directional locomotion, capability for instantaneous high-jerk acceleration ($> 3~\text{m/s}^2$), and stochastic trajectory evolution without explicit signaling.
    
    \item[\texttt{U02} Child:] Human individual with standing height between 80 cm and 120 cm, capable of lateral acceleration exceeding 2 m/s$^2$ within 0.3 seconds, and hand breadth less than 6 cm.
    \item[\texttt{U03} Mobility Impaired:] Human individual using a walker or cane for ambulation, requiring a minimum clear passage width of 90 cm, unable to step over obstacles taller than 2 cm.
    \item[\texttt{U08} Distracted:] User's gaze is fixed on a specific focal point with a visual field of view restricted to $<20$ degrees. Auditory perception is attenuated by external devices, raising the reaction threshold to auditory stimuli by 15--20 dB.
    \item[\texttt{U07} Drowsy:] Human individual exhibiting prolonged eyelid closure ($>1$ second), delayed reaction time ($>1.5$ s) to tactile stimuli, and reduced grip force ($<10$ N) during object holding.
    \item[\texttt{U05} Hearing Loss:] Human individual with audiometric thresholds exceeding 40 dB HL in the 500--4000 Hz frequency range.
    \item[\texttt{U10} Walking with Cane:] Human individual using a single-point cane, resulting in asymmetric gait cycle, average walking speed below 0.8 m/s, and constant unilateral ground contact.
\end{description}
\end{tcolorbox}

\subsection{Environment State Dimensions (E)}
\label{app:env_states}

These dimensions define the physical and sensory constraints of the workspace that interfere with robot perception or mobility.

\begin{tcolorbox}[colback=green!5, colframe=green!40, title=\textbf{Table B.2: Environment State Definitions}]
\footnotesize
\begin{description}[leftmargin=0.5cm, itemsep=2pt]
    \item[\texttt{E01} Low Light:] Ambient illuminance measured below 50 lux at floor level.
    \item[\texttt{E03} Wet Floor:] Floor surface covered with a continuous liquid layer, reducing static coefficient of friction to below 0.3.
    \item[\texttt{E04} Cluttered:] Primary walkway contains three or more unsecured objects on the floor within a 2-meter segment, reducing free path width to less than 70 cm.
    \item[\texttt{E05} Steam:] Airborne water vapor density sufficient to reduce optical transmission by more than 50\% at 850 nm wavelength over a 1-meter path.
    \item[\texttt{E06} Small Space:] Enclosed area with internal width between 60 cm and 120 cm, allowing only single-file passage between stationary obstacles.
    \item[\texttt{E08} Half-open Drawer:] Drawer extended 10--30 cm from its housing, presenting a flat obstacle at a height of 30--60 cm above floor level.
    \item[\texttt{E09} Stairs:] Vertical transition composed of steps with rise between 15 cm and 25 cm and no ramp or mechanical assistive device present.
    \item[\texttt{E10} Threshold:] Raised transition between adjacent floor surfaces with vertical height difference between 1.5 cm and 5 cm.
\end{description}
\end{tcolorbox}

\subsection{Supervision and Task Dimensions (S/T)}
\label{app:task_states}

\begin{tcolorbox}[colback=gray!10, colframe=gray!50, title=\textbf{Table B.3: Supervision and Task Definitions}]
\footnotesize
\begin{description}[leftmargin=0.5cm, itemsep=2pt]
    \item[\texttt{S01} Unsupervised:] No human agent is present in the environment capable of real-time observation or physical intervention.
    \item[\texttt{T01} Delivery:] The process of transferring an object to a designated recipient. Requires the robot's end effector to precisely reach a physically accessible zone and synchronize release timing with the recipient's grasp.
    \item[\texttt{T04} Emergency Response:] Task triggered by detection of a fall, vocal distress signal, or abrupt physiological change, requiring immediate action within 30 seconds.
    \item[\texttt{T03} Monitoring:] Task requiring continuous sensor lock on a target entity within a geofenced zone, updating position and posture data at a frequency greater than 5 Hz.
    \item[\texttt{T05} Repeated Request:] Task initiated by two or more identical user commands issued within a 10-second interval, indicating potential user urgency or system failure.
\end{description}
\end{tcolorbox}

\newpage
\section{Robot Functional Specifications} 
\label{app:robot_specs}

To ensure physical grounding, we defined a rigorous \textit{Robot Capability Profile} that is injected into all prompts. This profile specifies Testable thresholds (e.g., blind zones, latency) to prevent the LLM from hallucinating arbitrary capabilities.

\begin{tcolorbox}[colback=white, colframe=black, title=\textbf{Full Robot Specification Text}]
\scriptsize
\begin{verbatim}
[ROBOT FUNCTIONAL SPECIFICATIONS - STRICT PHYSICAL LIMITS]
(Use these limits to define verifiable thresholds. Do not assume capabilities not listed.)

 . PERCEPTION SYSTEMS (SENSING)
*   Head-Mounted Vision System:
    *   Function: Long-range detection, Semantic Classification, Body Keypoint Tracking.
    *   Physical Limits: Range 0.2m-5m. Blind Zone < 0.2m.
    *   Vertical Coverage: High vantage point; capable of seeing tabletops and floor.
*   Wrist-Mounted Vision System:
    *   Function: Close-range, occlusion-free manipulation guidance.
    *   Physical Limits: High precision. Blind Zone < 0.05m.
*   Chassis LiDAR System:
    *   Function: 360° Volumetric Obstacle Detection.
    *   Physical Limits: Range 0.1m-20m. Vertical Coverage: 0.05m to 3m.
    *   Constraint: Physically unable to detect overhanging obstacles above 3m.
*   Proximity Safety System:
    *   Function: Close-range blind-spot coverage & retraction guidance.
    *   Physical Limits: Range 0.05m-1.5m. 360° Sector Coverage.
*   Thermal Perception System:
    *   Function: Non-contact surface temperature monitoring.
    *   Capability: Identifies "Thermal Hazards" (e.g., hot plates, boiling liquids).
    *   Physical Limits: Effective Range 0.1m-1.2m. Accuracy +/- 2°C.
*   Localization & Proprioception System:
    *   Function: Real-time Map Pose, Room Identification, Kinematic State.
    *   Performance: Fuses IMU & Odometry for precise self-positioning and 
        Wheel Slip Detection (via cross-verification of velocity vectors).
*   Tactile & Force System:
    *   Function: Direct contact force (N), Torque (Nm), Slip detection.
    *   Performance: 1kHz reflex loop. Chassis Activation Threshold: > 5N.
*   System Memory & Logic (Compute Architecture):
    *   Safety Compute Engine (Dual-Loop Architecture):
        1.  Cognitive Loop (High-Level):
            *   Latency: 20Hz (~50ms).
            *   Constraint: Fast-moving objects (>2m/s) may traverse significant distance 
                between processing frames, causing "Ghosting".
        2.  Reflex Loop (Low-Level):
            *   Latency: 1kHz (<1ms).
            *   Constraint: Purely reactive. Zero predictive capability.

 . ACTUATION SYSTEMS (MOTION & MANIPULATION)
*   Mobility Chassis System:
    *   Function: Omni-directional (Holonomic) navigation.
    *   Capability: Simultaneous translation/rotation; Active Traction Control (ATC).
*   6-DOF Manipulation Arm:
    *   Function: Spatial positioning with Hybrid Position/Force Control.
    *   Capability: Impedance Control (Variable Stiffness).
    *   Safety Feature: Gravity Compensation.
*   Smart End-Effector (Gripper):
    *   Logic: Automatically modulates grip force based on tactile slip feedback.
    *   Safety Constraint: Supports Dynamic Force Ceiling.

 . CONTROL & SAFETY BEHAVIORS
*   Reactive Control Engine:
    *   Function: High-level task orchestration via Behavior Trees.
    *   Capability: Guarded Reach. Stops if external force exceeds safety threshold.
*   Retraction & Recovery Logic:
    *   Function: Safe exit from blind-spot collisions or stuck states.
    *   Input: Trajectory Buffer (Last 10 seconds).
    *   Action: Computes inverse vector to safely "back out".
\end{verbatim}
\end{tcolorbox}

\newpage
\section{Additional Visualizations for GPT-4o Backbone}
\label{app:gpt4o_viz}

In the main text, we primarily analyzed the results based on the DeepSeek-V3.2 backbone. To demonstrate the model-agnostic nature of our framework, we provide the corresponding visualizations for the \textbf{GPT-4o} backbone in this section.

\begin{figure}[h]
    \centering
    \begin{subfigure}[b]{0.48\textwidth}
        \centering
        \includegraphics[width=\textwidth]{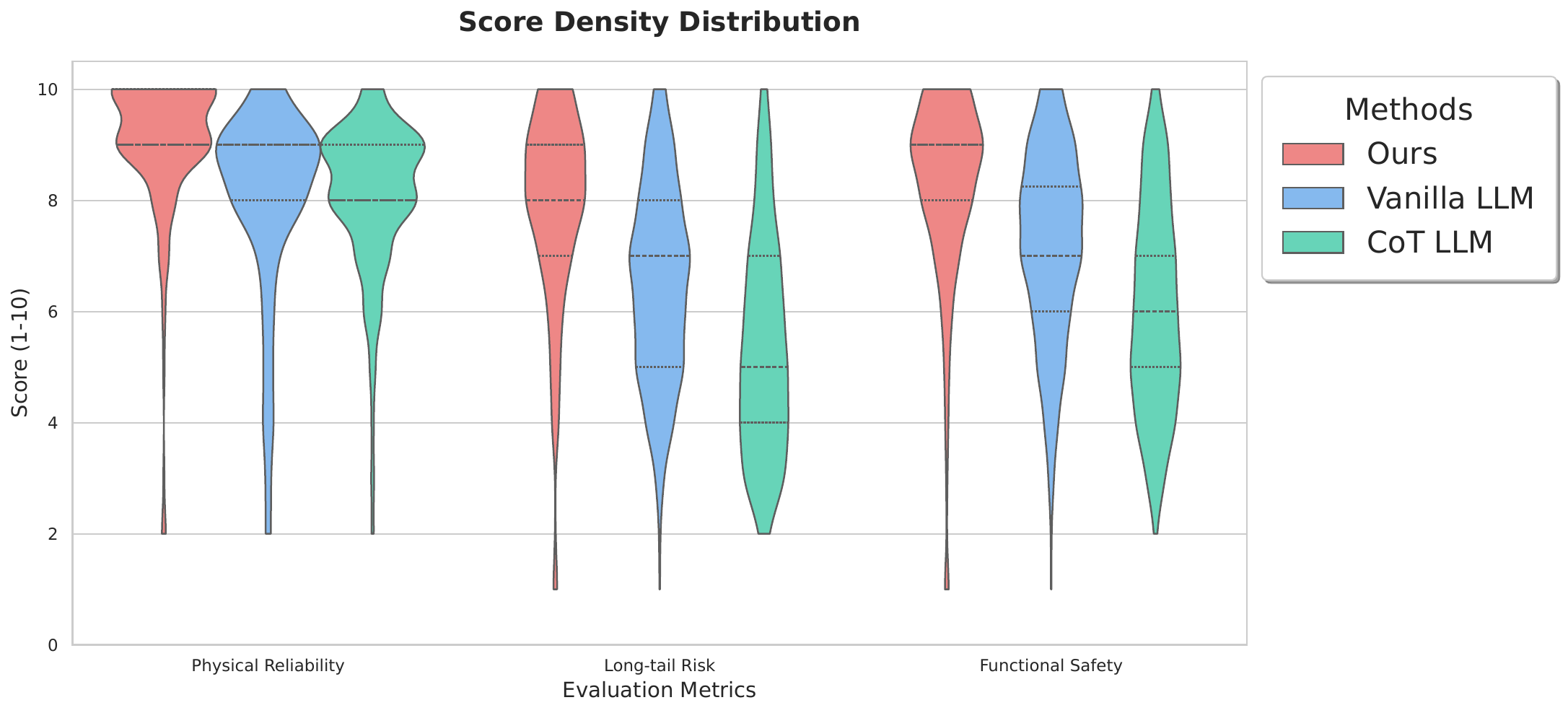}
        \caption{Score Distribution (Violin Plot)}
        \label{fig:violin_gpt}
    \end{subfigure}
    \hfill
    \begin{subfigure}[b]{0.48\textwidth}
        \centering
        \includegraphics[width=\textwidth]{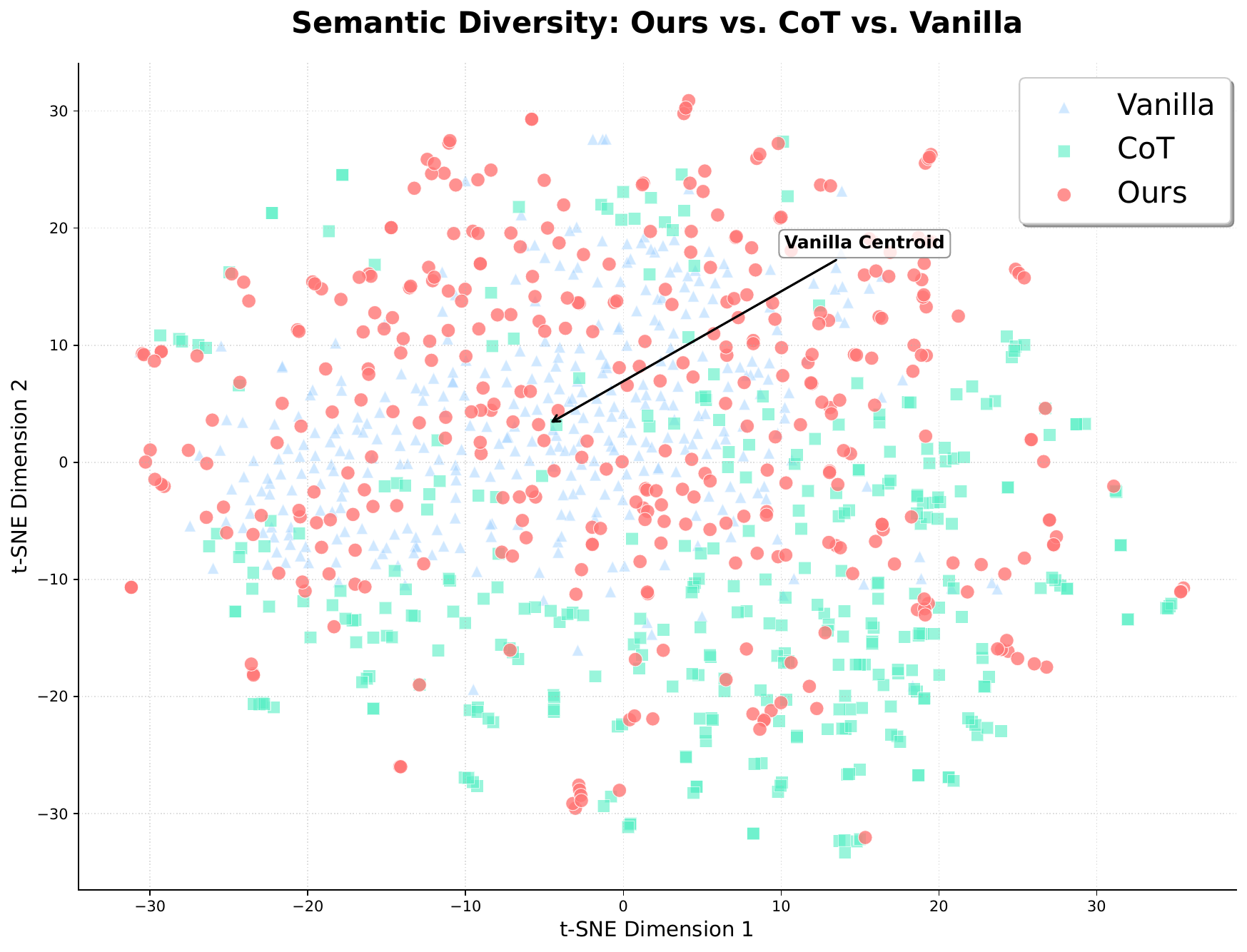}
        \caption{Feature Space Visualization (t-SNE)}
        \label{fig:tsne_gpt}
    \end{subfigure}
    
    \caption{\textbf{Visualizations for the GPT-4o Backbone.} (a) The violin plot confirms that our method maintains a ``top-heavy'' high-quality distribution even on the stronger GPT-4o model, whereas baselines still exhibit long-tail risks. (b) The t-SNE projection shows that our method (red) covers a distinct and broader semantic space compared to the baselines, consistent with the findings on DeepSeek-V3.2.}
    \label{fig:gpt_viz_combined}
\end{figure}
\newpage
\section{Prompt Construction \& Templates}
\label{app:prompt_templates}

In this section, we detail the exact prompt templates used for the evaluation phase of our framework.

\subsection{Functional Safety Requirement (FSR) Audit Prompt}
\label{app:fsr_audit_prompt}

\begin{tcolorbox}[colback=green!5, colframe=green!40, title=\textbf{Template: FSR Technical Audit}, breakable]
\begin{lstlisting}[basicstyle=\scriptsize\ttfamily, breaklines=true, columns=fullflexible, xleftmargin=0pt, extendedchars=true]
EVALUATION_PROMPT_TEMPLATE = """You are a Principal Functional Safety Auditor and Robotics Systems Architect. Your task is to perform a rigorous, standalone evaluation of a set of Functional Safety Requirements (FSRs) generated for a specific robotic scenario.

[IMPORTANT INSTRUCTION]
Ignore the name/label of the method (e.g., "Method A and FSR ID"). Focus exclusively on the technical quality, completeness, and hardware-grounding of the FSRs provided.

[INPUT DATA]
Seed Scenario: {seed_scenario}

Target FSRs to Evaluate:
{fsr_content}

[STAGE 1: AUDIT FOUNDATIONS - THE LAWS OF PHYSICS AND HARDWARE]
[ROBOT FUNCTIONAL SPECIFICATIONS - EXTERNAL REFERENCE]
**Refer to the external technical document: {{ROBOT_HARDWARE_SPECIFICATION_DOC}}** 
(This document defines the absolute physical constraints, including Perception Systems, Actuation Systems, Interaction Systems, and Control/Safety Behaviors. Any FSR violating these limits is considered a "Hallucination" and must be severely penalized.)

[STAGE 2: SCORING RUBRICS (THE AUDIT STANDARD)]

Metric 1: Capability Compliance & Grounding (CC)
*   9-10 (Hardware-Optimized): The method perfectly leverages specific hardware capabilities (e.g., specific control modes, memory engines, sensor fusion) to solve safety problems. It explicitly accounts for sensor limits and blind spots defined in the spec.
*   7-8 (Advanced Adaptation): The method aligns well with hardware specs and accounts for most sensor limitations, though it may not fully exploit advanced features like specific memory engines or complex fusion algorithms.
*   5-6 (Generic): The method uses generic logic without leveraging the robot's specific advanced capabilities. It does not violate limits but does not optimize for them.
*   3-4 (Suboptimal Alignment): The method has loose integration with hardware specs. It fails to account for obvious sensor limits or blind spots, potentially leading to unstable performance in specific hardware environments.
*   1-2 (Hallucination/Violation): The method demands capabilities the robot does not have. Specifically, it requires detection ranges or sensor modalities that contradict the [ROBOT FUNCTIONAL SPECIFICATIONS]. This is an automatic failure.

Metric 2: Scenario Risk Coverage (PRC)
*   9-10 (Exhaustive Coverage): The FSRs identify and mitigate all primary hazards, secondary consequences (e.g., inertia, load stability), and **long-tail/edge-case risks** specific to this scenario (e.g., rare environmental interferences, complex human behaviors, or multi-system failures).
*   7-8 (Advanced Physics & Risk): The method goes beyond simple collision avoidance and addresses dynamic effects and most secondary risks, but coverage of complex long-tail interaction dynamics is incomplete.
*   5-6 (First-Order/Basic): The method covers direct risks only. It addresses primary collisions and slips but ignores the secondary effects of the robot's reaction or scenario-specific nuances.
*   3-4 (Basic Physics): The method has a narrow understanding of physical risks, identifying only the most obvious contact risks while neglecting friction, slippage, or basic kinematic constraints.
*   1-2 (Superficial): The method relies solely on semantic rules and ignores physical dynamics and scenario-specific risks entirely.

Metric 3: Logic Robustness & Continuity (LRC)
*   9-10 (Closed-Loop System): The method defines clear Entry AND Exit conditions for every safety state. It utilizes recovery logic (e.g., Trajectory Buffer) to exit failure modes instead of just freezing. It uses memory/persistence to handle perception gaps.
*   7-8 (Robust Closed-Loop): The method defines clear entry/exit conditions and includes basic recovery logic. However, it may be slightly lacking in handling complex perception gaps or long-term memory persistence.
*   5-6 (Open-Loop): The method defines when to stop but provides vague or missing conditions for when to resume operation.
*   3-4 (Fragmented Logic): The method has basic trigger logic, but state transitions are inconsistent and lack clear recovery paths, likely leading to frequent freezing or requiring manual intervention.
*   1-2 (Deadlock/Dangerous): The method creates logical deadlocks (robot freezes permanently) or prescribes dangerous actions that violate basic safety principles.

[STAGE 3: THE AUDIT PROCESS (THINK STEP-BY-STEP)]

**STEP 1: INDEPENDENT HARDWARE REALITY CHECK**
*   Cross-reference every FSR against the [ROBOT FUNCTIONAL SPECIFICATIONS].
*   Identify any requirement that assumes a sensor range, coverage, or capability that is explicitly listed as a limitation or not listed at all.

**STEP 2: SCENARIO-SPECIFIC HAZARD ENUMERATION**
*   Analyze the "Seed Scenario" deeply. List all potential risks:
    1. Primary Risks (e.g., direct collision).
    2. Secondary Risks (e.g., inertia after emergency stop, object dropping).
    3. **Long-tail Risks** (e.g., sensor blinding by sunlight, floor slip during heavy load, unexpected human interference).
*   Check if the provided FSRs cover these specific points.

**STEP 3: LOGICAL STATE MACHINE & RECOVERY ANALYSIS**
*   Analyze the "Trigger -> Action -> Recovery" loop.
*   Does the method utilize the **Trajectory Buffer** or **Object Persistence Engine** to handle blind spots or recovery?

**STEP 4: FINAL SCORING**
*   Assign scores based on the evidence. Be extremely critical of generic requirements that ignore the robot's advanced hardware.

[OUTPUT FORMAT]

**1. Detailed Audit Analysis (Chain of Thought)**
*   **Hardware Alignment Analysis**: [Detailed reasoning on hardware alignment]
*   **Scenario Risk & Long-tail Coverage Analysis**: [Detailed reasoning on how well the FSRs cover primary, secondary, and long-tail risks of the specific scenario]
*   **Logic Robustness & Recovery Analysis**: [Detailed reasoning on state transitions and recovery logic]

**2. Capability Violation Report**
*   Identify and list any specific FSR IDs that contradict the [ROBOT FUNCTIONAL SPECIFICATIONS]. If none, state "None".

**3. Final Scores**

 **Capability Compliance (CC)** | X/10 | [Brief justification] 
 **Scenario Risk Coverage (PRC)** | X/10 | [Brief justification]
|**Logic Robustness (LRC)** | X/10 | [Brief justification] 

**4. Final Verdict**
*   [Summary of the FSR quality ]
"""
\end{lstlisting}
\end{tcolorbox}

\subsection{Scenario Quality Evaluation Prompt}
\label{app:scenario_eval_prompt}

\begin{tcolorbox}[colback=blue!5, colframe=blue!40, title=\textbf{Template: Scenario Quality Evaluation}, breakable]
\begin{lstlisting}[basicstyle=\scriptsize\ttfamily, breaklines=true, columns=fullflexible, xleftmargin=0pt, extendedchars=true]
EVALUATION_PROMPT_TEMPLATE = """You are a robotic safety engineer. 

[REFERENCE: ROBOT CAPABILITIES & CONSTRAINTS]
**Refer to the external technical document: {{ROBOT_HARDWARE_SPECIFICATION_DOC}}**
(All evaluations must be grounded in the specific hardware and functional limits of the robot platform.)

I will provide three sets of scenarios (Method A, Method B, and Method C), all generated from the same seed scenario.

Please evaluate **each scenario independently**, and assign a quantitative score (1-10 points) based on the strict abstract metrics defined below.

**STEP 1: Global Comparative Thinking**
Before any scoring, perform a high-level comparative analysis of Method A, B, and C. Address the following:
- **Methodological Divergence**: How does each method approach the "expansion" of the seed? (e.g., semantic changes vs. physical perturbations).
- **Constraint Adherence**: Which method stays truest to the seed's environment, and which one tends to "hallucinate" external factors?
- **Risk Profile**: Compare the "sophistication" of the risks discovered. Are they identifying simple failures or complex, multi-factor safety boundary violations?
- **FSR Derivation Potential**: Which method provides the most complete causal chain (Trigger -> System Failure -> Hazard) to support the derivation of Functional Safety Requirements?

**STEP 2: Individual Scenario Scoring**
Evaluate each scenario based on the metrics below, informed by your global analysis.


### Scoring Metrics (1-10 points)

**1. Physical Reliability (Higher is Better)**
*   **Definition**: Whether the scenario adheres to real-world physics AND strictly maintains the "closed-world" constraints of the seed scenario.
*   **High Score (8-10)**: The scenario operates **strictly** using only the entities, agents, and environmental features explicitly defined or inherently implied in the seed.
*   **Medium Score (4-7)**: Physical interactions are valid, but the scenario makes **minor assumptions** about environmental states without introducing new active agents.
*   **Low Score (1-3)**: **CRITICAL FAILURE**. The scenario introduces **new active entities, obstacles, or external forces** not present in the seed.

**2. Long-tail Risk Discovery Capability (Higher is Better)**
*   **Definition**: The degree to which the scenario uncovers statistically rare, concealed, or system-boundary hazards *within the bounds of the seed context*.
*   **High Score (8-10)**: Identifies risks characterized by **multi-factor coupling**, **sensor/actuator physical limits**, or **semantic ambiguity**.
*   **Medium Score (4-7)**: Risks are valid but represent standard operational hazards or "Fat-tail" events.
*   **Low Score (1-3)**: Describes routine operations with no significant hazard; risks are trivial.

**3. Functional Safety Requirement Derivation Capability (Higher is Better)**

*   **Definition**: The degree to which the scenario clarifies the failure mechanism (the "Why" and "How"), enabling the systematic synthesis of Functional Safety Requirements (FSR).
*   **High Score (8-10)**: The scenario explicitly identifies the triggering condition and the system performance limit. The causal chain (Trigger -> System Behavior -> Hazard) is logically complete, making the derivation of quantitative, verifiable safety requirements straightforward.
*   **Medium Score (4-7)**: The scenario identifies a credible hazard and its qualitative root cause. It supports the definition of high-level Safety Goals (SG), but the logic lacks the specific parameters needed to synthesize precise FSRs without further decomposition.
*   **Low Score (1-3)**: CRITICAL FAILURE. The scenario describes a "bad outcome" without explaining the underlying failure logic or triggering events. It offers no actionable path for safety requirement engineering.
**Please reply in the following format**:

[Global Comparative Analysis]
(Provide your deep-dive comparison here, identifying the "DNA" of each method's approach.)


[Method A Scenario Evaluation]
Scenario 1: [Brief justification]  
Physical Reliability: X pts, Long-tail Risk: X pts, Safety Requirements: X pts
...

[Method B Scenario Evaluation]
Scenario 1: [Brief justification]  
Physical Reliability: X pts, Long-tail Risk: X pts, Safety Requirements: X pts
...

[Method C Scenario Evaluation]
Scenario 1: [Brief justification]  
Physical Reliability: X pts, Long-tail Risk: X pts, Safety Requirements: X pts
...

[Conclusion]  
A brief conclusion comparing the three methods.

**Seed Scenario**:  
{seed_scenario}

**Method A Scenarios**:  
{method_a_scenarios}

**Method B Scenarios**:  
{method_b_scenarios}

**Method C Scenarios**:  
{method_c_scenarios}

Please begin your evaluation:"""
\end{lstlisting}
\end{tcolorbox}

\end{document}